\documentclass[11pt]{article}

\usepackage[preprint]{acl}

\usepackage{times}
\usepackage{latexsym}

\usepackage[T1]{fontenc}

\usepackage[utf8]{inputenc}

\usepackage{microtype}

\usepackage{inconsolata}

\usepackage{graphicx}

\usepackage[whole]{bxcjkjatype} 
\usepackage{hyperref}
\usepackage{url}
\usepackage{booktabs}
\usepackage[utf8]{inputenc}
\usepackage{inconsolata}
\usepackage{amsmath,amssymb}
\usepackage{graphicx,xcolor}  
\usepackage{subcaption}
\usepackage{multirow}
\usepackage{makecell}
\usepackage{mathtools}
\usepackage{bm}
\usepackage{tikz}
\usepackage{tabularx}
\usepackage{pifont}
\usepackage[most]{tcolorbox}
\usepackage{fontawesome5} 

\usepackage{lineno}

\definecolor{darkblue}{rgb}{0, 0, 0.5}
\hypersetup{colorlinks=true, citecolor=darkblue, linkcolor=darkblue, urlcolor=darkblue}

\definecolor{beforebg}{HTML}{DBEAFE}
\definecolor{beforefg}{HTML}{1E40AF}
\definecolor{targetbg}{HTML}{F3F4F6}
\definecolor{targetfg}{HTML}{374151}
\definecolor{successbg}{HTML}{ECFDF5}
\definecolor{successfg}{HTML}{065F46}
\definecolor{successborder}{HTML}{A7F3D0}
\definecolor{failbg}{HTML}{FEF2F2}
\definecolor{failfg}{HTML}{991B1B}
\definecolor{failborder}{HTML}{FECACA}
\definecolor{successlabel}{HTML}{059669}
\definecolor{faillabel}{HTML}{DC2626}

\newcommand{\rel}[2]{\emph{#1 $\subset$ #2}}

\title{Linear Representations of Hierarchical Concepts in Language Models}

\author{
Masaki\,Sakata$^{1,2}$\hspace{1em}
\textbf{Takumi\,Ito}$^{1,4}$\hspace{1em}
Sho\,Yokoi$^{3,1,2}$ \\ 
\textbf{Kentaro\,Inui}$^{5,1,2}$\hspace{1em}
\textbf{Benjamin\,Heinzerling}$^{2,1}$
\\[2pt]
$^{1}$ Tohoku University\hspace{1em}
$^{2}$ RIKEN\hspace{1em}
$^{3}$ NINJAL\hspace{1em}
$^{4}$ Langsmith Inc.\hspace{1em}
$^{5}$ MBZUAI\hspace{1em}
\\
\texttt{sakata.masaki.s5@dc.tohoku.ac.jp}\hspace{1em}
\texttt{t-ito@tohoku.ac.jp}\hspace{1em}
\texttt{yokoi@ninjal.ac.jp}\\
\texttt{kentaro.inui@mbzuai.ac.ae} \hspace{1em} 
\texttt{benjamin.heinzerling@riken.jp} \\
}

\begin{document}
\maketitle
\begin{abstract}
Prior work on hierarchical representations in language models (LMs) has focused on static unembedding representations and single-token concepts.
To examine hierarchical representations across intermediate layers and for multi-token entities, we extend the analytical approach of \citet{lrc} to entire concept hierarchies.
We refer to this analytical setup as \emph{Linear Hierarchical Encoding} (LHE).
Across four LMs and five domains (locations, persons, organizations, organisms, and research topics), parent--child relations are linearly decodable, and interventions along the extracted directions can causally steer predictions.
The behaviorally relevant information is concentrated in relatively low-dimensional, domain-specific subspaces, yet their internal organization is similar across domains.
Finally, editing only a direct-parent relation can produce hierarchy-consistent changes and preservation effects in predictions about higher ancestors, although success varies across domains and models.
These results show that LMs instantiate concept hierarchies in distinct subspaces with similar geometry, and that these representations can influence hierarchical predictions.~\faGithub~\href{https://github.com/masaki-sakata/linear-hierarchical-encoding}{\nolinkurl{github.com/masaki-sakata/
linear-hierarchical-encoding}}
\end{abstract}

\section{Introduction}
\label{sec:introduction}

Whether and how language models (LMs) form geometric representations of concepts~\citep{gardenfors2004conceptual} is of both theoretical~\citep{huh2024position,mollo2026vector} and practical importance~\citep{zou2023representation,templeton2024scaling,arditi2024refusal}.
Recent work has answered this question positively by identifying geometric representations of concept structures ranging from  colors~\citep{abdou-etal-2021-language} and relational knowledge~\citep{merullo-etal-2024-language} to geographic location and time~\citep{repr_space,heinzerling-inui-2024-monotonic}.
Most recently, \citet{geo_hierarchical_concepts} extended this line of research to concept hierarchies, showing that the unembedding layer encodes hierarchical relations geometrically.

\begin{figure*}[t]
\centering
\includegraphics[width=\linewidth]{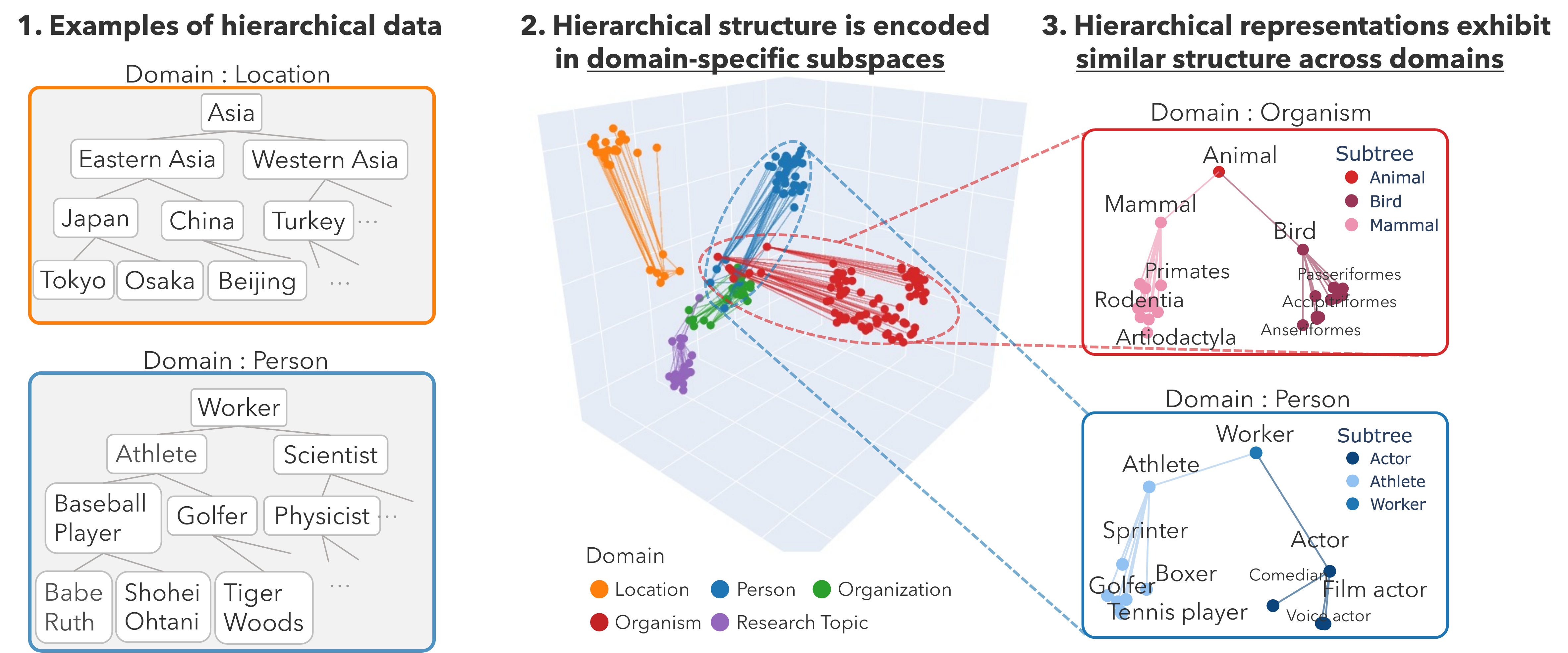}
\caption{Overview of our representational findings. LMs represent hierarchical concepts (left) in domain-specific subspaces of intermediate-layer activations (center), while the organization within these subspaces is similar across domains (right). The middle and right panels visualize, using PCA, the representations obtained by applying a linear transformation trained on hierarchical data to the hidden-state vectors corresponding to each concept in intermediate layers of Llama 3.1 8B. See App.~\ref{sec:appendix_pca} for details.}
\label{fig:pontie}
\end{figure*}

However, understanding of how LMs represent hierarchical concepts is still limited.
First, \citet{geo_hierarchical_concepts} only study static representations in the unembedding layer. Since LMs encode many relations across intermediate layers~\citep{merullo-etal-2024-language,lre,lrc}, any analysis restricted to a single layer captures only a small part of the model's representational capacity.
Second, \citeauthor{geo_hierarchical_concepts}'s method is limited to concepts that correspond to a single token in the LM's vocabulary and hence cannot handle hierarchical relations involving multi-token entities, such as \rel{Babe Ruth}{Baseball Player}.

\begin{figure*}[t]
\centering
\includegraphics[width=0.8\linewidth]{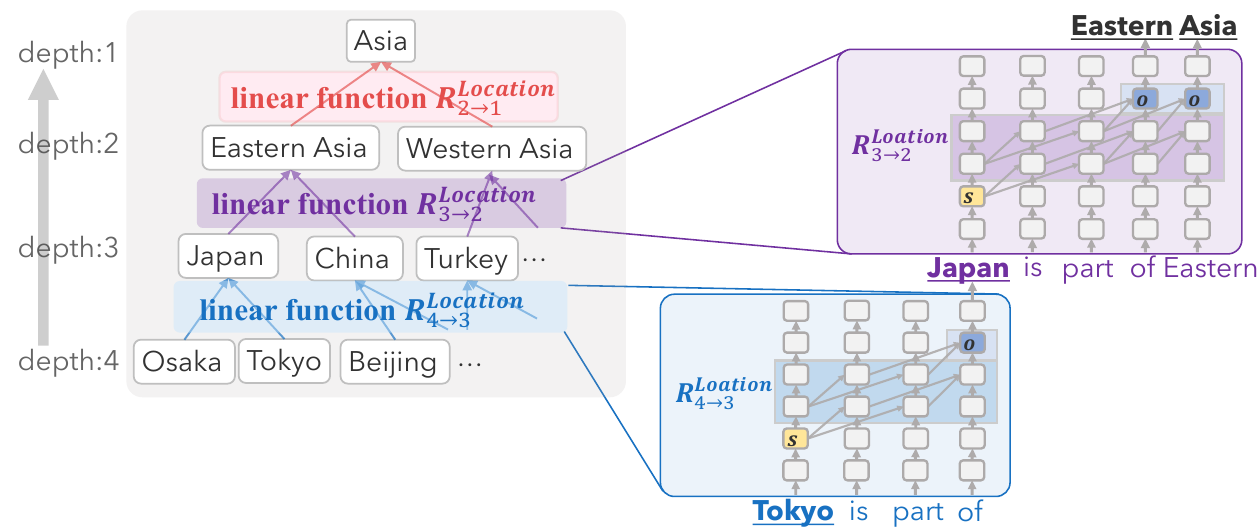}
\caption{Overview of \emph{Linear Hierarchical Encoding} (LHE). 
Given a hierarchical relation, LHE learns depth-specific and domain-specific linear transformations that map a child concept representation to the representation of its parent concept in LM intermediate layers. 
}
\label{fig:method_overview}
\end{figure*}

We address this gap with \emph{Linear Hierarchical Encoding} (LHE), a framework for analyzing cross-layer hierarchy representations involving diverse, entity-centric concepts.
Building on Linear Relational Concepts (LRC)~\citep{lrc}, LHE maps a child's representation at an earlier layer to its parent's representation at a later layer (Fig.~\ref{fig:method_overview}).
For example, in the context ``Tokyo is part of,'' it learns a linear transformation from the intermediate-layer representation of ``Tokyo'' to that of ``Japan.''
We evaluate LHE on concept hierarchies with multi-token entities (Fig.~\ref{fig:pontie}, left).

Our study is organized around three research questions:
\textbf{RQ1:} Are hierarchical relations linearly encoded and behaviorally relevant?
\textbf{RQ2:} How is this information organized across dimensions, domains, and hierarchy depths?
\textbf{RQ3:} Do local edits produce hierarchy-consistent effects on related ancestor predictions?
These questions separate evidence that a method can recover and steer parent relations from the representational geometry revealed by that method, and from the stronger causal test of whether an edit coheres with a broader hierarchy.

We make three corresponding contributions.
First, we show across four LMs and five domains that parent--child relations are linearly decodable and that LHE directions can causally steer predictions; we also identify cases where linear decodability and causal use dissociate.
Second, we show that behaviorally relevant hierarchy information occupies relatively low-dimensional, domain-specific subspaces, while the internal organization of those subspaces is similar across domains.
Third, we find that editing only a direct-parent relation can produce partially consistent Change and Preserve effects on predictions about higher ancestors.
Together, these findings suggest that LMs do not store hierarchies in one universal coordinate system; rather, distinct domain-specific subspaces instantiate a shared geometric organization that can influence hierarchical predictions.

\section{Related Work}
\label{sec:related_work}

\subsection{Hierarchical Representations}
Prior work examines hierarchy at distinct linguistic levels and for different purposes.
\citet{hewitt-manning-2019-structural} and \citet{chen2021probing} study the geometric encoding of syntactic dependency structure in contextual word representations.
\citet{liu-neubig-2022-representations} examine phrase compositionality by asking whether representations of whole phrases can be predicted from those of their constituents.
These studies focus on syntactic structure and phrase compositionality rather than entity-centric concept hierarchies.
Work on lexical-semantic hierarchies instead evaluates whether encoder representations satisfy intrinsic taxonomic properties~\citep{LM_Hierarchy_Properties} or retrains encoders to represent taxonomies in hyperbolic space~\citep{NEURIPS2024_1a970a3e}.
More directly related work examines how hierarchical relations among concepts are reflected in representation geometry.
\citet{Hierarchical-sae-2025} use a synthetic hierarchical generative model to test whether SAEs recover hierarchical features.
\citet{geo_hierarchical_concepts} estimate concept vectors from LM unembeddings using WordNet.
They theoretically derive and empirically validate geometric properties of hierarchical relations, including orthogonality between a superordinate concept direction and directions distinguishing its subordinate categories.
Whereas their analysis relies on unembedding representations of individual vocabulary tokens, we examine concept hierarchies across intermediate layers of decoder LMs (Fig.~\ref{fig:method_overview}) and extend the scope to multi-token entities and class names (Fig.~\ref{fig:pontie}, left).

\subsection{Linear Relational Representations and Interventions}
Under the linear representation hypothesis, a growing body of work has found geometric structure for concepts and relations, including color, sentiment, knowledge, space, and time~\citep{abdou-etal-2021-language,nanda-etal-2023-emergent,Sentiment-2023,Knowledge-direction-2023,repr_space,heinzerling-inui-2024-monotonic}.
From a methodological perspective, the work most closely related to ours is that on Linear Relational Embeddings (LRE)~\citep{lre} and Linear Relational Concepts (LRC)~\citep{lrc}.
LRE considers knowledge represented as subject--relation--object triples, such as (Miles Davis, plays, trumpet), and linearly approximates the computation performed in an LM's intermediate layers when predicting the object from the subject and relation.
It shows that this computation can be approximated reasonably well by a single linear transformation.
LRC extends LRE to the multi-token setting and shows that it can identify directions in intermediate-layer representations that correspond to concepts.
Our LHE builds on the LRC framework by learning separate models for each hierarchical depth and domain.
This allows us to study how hierarchical information is encoded in LM intermediate-layer representations in relation to the computations that produce those representations.

\section{Linear Hierarchical Encoding}

\subsection{Task Formulation}
We represent each instance on a hierarchy $\mathcal{T}$ as a triple $(s,r,o)$, where the subject $s$ is a child node and the object $o$ is its parent node. The relation $r$ is a label indicating a hierarchical transition, and we distinguish $r$ by depth. 
For example, in the chain Japan$\rightarrow$Eastern Asia$\rightarrow$Asia,
the transition between depths 2 and 3 is represented as
$(s,r,o)=(\text{Japan},\textsc{part-of}_{2\text{--}3},\text{Eastern Asia})$.

\subsection{Estimating Child-to-Parent Transformations}
In plain terms, LHE maps the representation of a child entity at an earlier layer to the representation of its parent label at a later layer. Its low-rank inverse then converts parent representations into directions in the child-representation space.

Following \citet{lrc}, we represent the child in example $n$ by the hidden state at its last token position $i$ in the subject layer $\ell_s$, i.e., $\mathbf{s}_n=\mathbf{h}^{(\ell_s)}_i$.
Let $\mathcal{C}_n$ denote the prompt context for example $n$.
Given a child state $\mathbf{s}$ at that position, $F(\mathbf{s},\mathcal{C}_n)$ runs the LM in that context and returns the mean hidden state over the parent tokens at the object layer $\ell_o$.
The parent representation for example $n$ is $\mathbf{o}_n=F(\mathbf{s}_n,\mathcal{C}_n)$.

Following LRE~\citep{lre}, we approximate this mapping for each relation $r$ by $R_r(\mathbf{s})=\mathbf{W}_r\mathbf{s}+\mathbf{b}_r$.
For each training example $n$ of relation $r$, we linearize $F$ around $\mathbf{s}_n$ with $\mathcal{C}_n$ held fixed.
We average the resulting Jacobians $\left.\frac{\partial F(\mathbf{s},\mathcal{C}_n)}{\partial \mathbf{s}}\right|_{\mathbf{s}=\mathbf{s}_n}$ and their corresponding bias terms to obtain $\mathbf{W}_r$ and $\mathbf{b}_r$, respectively.
Following \citet{lrc}, we use a low-rank pseudo-inverse $\mathbf{W}_r^{\dagger}$ to define the inverse mapping
$R_r^{\dagger}(\mathbf{o})=\mathbf{W}_r^{\dagger}(\mathbf{o}-\mathbf{b}_r).$
We apply this inverse map to obtain a concept vector $\mathbf{v}_{r,c}$ for each parent concept $c$.

\subsection{Deriving Parent Concept Directions}
Intuitively, $\mathbf{v}_{r,c}$ can be seen as a prototype direction in the child-representation space for nodes with parent concept $c$.
For relation $r$, let $\mathcal{I}_{r,c}=\{n \mid o_n=c\}$ be the index set of examples whose parent is $c$.
For each such example, we compute $\mathbf{o}_n$ in the prompt context of its own child and apply the inverse map.
We average these mapped representations for parent concept $c$ and normalize the result to obtain $\mathbf{v}_{r,c}$.
\begin{align*}
\tilde{\mathbf{v}}_{r,c}&=\frac{1}{|\mathcal{I}_{r,c}|}\sum_{n\in\mathcal{I}_{r,c}}
\mathbf{W}_r^{\dagger}(\mathbf{o}_n-\mathbf{b}_r), \\
\mathbf{v}_{r,c}&=\frac{\tilde{\mathbf{v}}_{r,c}}{\|\tilde{\mathbf{v}}_{r,c}\|_2}.
\end{align*}
This yields concept vectors $\mathbf{v}_{r,c}$ corresponding to concepts such as ``part-of:Japan'' and ``part-of:Asia''. For example, the ``part-of:Asia'' vector serves as a prototype for child-node representations belonging to ``Asia''.
For each hierarchical relation $r$, we compute the set $\{\mathbf{v}_{r,c}\}_c$; this procedure defines \emph{Linear Hierarchical Encoding (LHE)}.

\subsection{Map Granularities}
Hierarchies contain the same relation at different depths and in semantically different domains. By default, LHE therefore indexes $r$ by both domain and depth transition and estimates one map per domain--depth pair (Fig.~\ref{fig:method_overview}).
Throughout the paper, \emph{LHE} without further qualification refers to this default configuration.
Sec.~\ref{sec:domain_specificity} compares this default with maps shared across depths, domains, or both.

\section{Experimental Setup}
\label{sec:Experiments_setup}

\paragraph{Generalization protocol.}
To evaluate generalization beyond memorized training subtrees, we partition each domain tree at the root level.
For example, in Locations, we train on subtrees under Europe, the Americas, and Oceania and evaluate on those under Asia and Africa.
For LHE, we split the training partition into train and dev and select each model's hyperparameters by maximizing the mean of Accuracy and Causality on dev.
We then refit LHE on the combined train and dev splits and evaluate it on the held-out test subtrees (App.~\ref{sec:appendix_hyperparameter}).

\paragraph{Evaluation.}
We use two complementary metrics. \textit{Accuracy} measures linear decodability, namely, whether the correct parent can be recovered linearly from a child's internal representation using LHE concept vectors. \textit{Causality} measures whether the extracted directions affect model behavior.
Given a test-time child representation $\mathbf{a}$, we predict the parent whose concept vector has the largest inner product:
$\hat{y}=\arg\max_{c}\ \mathbf{v}_{r,c}\cdot\mathbf{a}.$
Accuracy is the proportion of test examples for which $\hat{y}$ matches the gold parent.

For Causality, we test whether adding and subtracting concept vectors changes prediction probabilities in the intended direction.
For example, given ``Paris is part of'', we subtract ``part of: France'' and add ``part of: Germany'' to the representation of ``Paris'', then test whether Germany scores higher than France.
When changing the original parent $c$ to a target $c'$, we intervene on the final-token child representation at every layer as
\[
\mathbf{h}^{(\ell)}_i \leftarrow \mathbf{h}^{(\ell)}_i + \beta\|\mathbf{h}^{(\ell)}_i\|_2(\mathbf{v}_{r,c'}-\mathbf{v}_{r,c}).
\]
This all-layer protocol follows \citet{lrc} and related work on multilayer activation interventions~\citep{arditi2024refusal,pmlr-v235-liu24bx,wang2025semantics}.
\footnote{
We report an ablation study of the intervention layers in App.~\ref{sec:appendix_intervention_layers}.
}
An intervention succeeds if the predicted probability of $c'$ exceeds that of $c$ after editing.
For multi-token parents, we use the minimum token probability over the parent tokens, following \citet{lrc}.
Strong performance on both metrics indicates that hierarchy information is not only decodable but also behaviorally relevant.

\paragraph{Data.}
To study hierarchical relations in entity-centric concept sets, we use five domain-specific hierarchical datasets spanning locations~\citep{repr_space, lukes_iso3166_regional_codes_v10}, persons, organizations~\citep{wikidata}, organisms~\citep{geo_hierarchical_concepts}, and research topics~\citep{openalex}.
These datasets contain diverse multi-token entities with each node having a single parent.
Dataset statistics are reported in Tbl.~\ref{tab:domain-depth-counts}, and concrete examples are provided in Fig.~\ref{fig:pontie} (left) and Fig.~\ref{fig:data_example}. See App.~\ref{sec:appendix_data} for details.
To focus our analysis on cases where the LM is likely to already possess knowledge of the hierarchical relation, we filter the dataset to instances that the model answers correctly in a few-shot multiple-choice question answering setting.
Statistics for the filtered data for each LM are reported in Tbl.~\ref{tab:domain-depth-counts-all-models}.
We use the same prompt template both for this filtering step and for all experiments; the prompt is shown in Fig.~\ref{fig:prompt}.

\begin{table}[t]
\centering
\small
\begin{tabular}{lrrrrr}
\toprule
Depth & Loc. & Top. & Per. & Org. & Ogm. \\
\midrule
0 (Root) & 1 & 1 & 1 & 1 & 1 \\
1 & 5 & 4 & 6 & 6 & 6 \\
2 & 22 & 25 & 35 & 21 & 87 \\
3 & 263 & 219 & 224 & 235 & 461 \\
4 & 27971 & 4464 & -- & -- & -- \\
\bottomrule
\end{tabular}
\caption{Number of examples by hierarchical depth across domains. Loc., Top., Per., Org., and Ogm. denote Locations, Research Topics, Persons, Organizations, and Organisms, respectively.}
\label{tab:domain-depth-counts}
\end{table}

\paragraph{LMs.}
We analyze four decoder-only LMs: Llama 3.2 3B~\citep{llama3_2_model_card}, Llama 3.1 8B~\citep{llama3_1}, Qwen3 8B, and Qwen3 14B~\citep{qwen3}.
The details of each model are summarized in Tbl.~\ref{tab:models_detail}.

\begin{table}[t]
\centering
\small
\begin{tabular}{lcc}
\toprule
Method & Accuracy & Causality \\
\midrule
Input averaging & 0.40 & 0.16 \\
SVM             & 0.55 & 0.35 \\
LHE             & \textbf{0.75} & \textbf{0.56} \\
\bottomrule
\end{tabular}
\caption{Comparison of LHE with linear baselines on Llama 3.1 8B. LHE denotes the default configuration with one map per domain--depth pair. Scores are macro-averaged over the five domains, and all rows use the same model-filtered test data. Results for all four models are reported in Tbl.~\ref{tab:acc_causality_all_models} in App.~\ref{sec:appendix_other_model_results}.}
\label{tab:acc_causality}
\end{table}

\begin{table}[t]
\centering
\small
\begin{tabular}{lccccc}
\toprule
Model & Loc. & Top. & Per. & Org. & Ogm. \\
\midrule
\multicolumn{6}{c}{Accuracy (LHE)} \\
\cmidrule(lr){1-6}
Llama 3.2 3B & 0.76 & \textbf{0.69} & \textbf{0.92} & 0.96 & \textbf{0.90} \\
Llama 3.1 8B & 0.70 & 0.53 & 0.88 & 0.96 & 0.75 \\
Qwen3 8B     & \textbf{0.77} & 0.62 & \textbf{0.92} & \textbf{0.98} & 0.81 \\
Qwen3 14B    & 0.76 & 0.58 & 0.85 & 0.90 & 0.80 \\
\midrule
\multicolumn{6}{c}{Causality (LHE)} \\
\cmidrule(lr){1-6}
Llama 3.2 3B & 0.63 & 0.14 & \textbf{0.69} & \textbf{0.75} & \textbf{0.65} \\
Llama 3.1 8B & \textbf{0.65} & 0.12 & 0.64 & 0.52 & 0.53 \\
Qwen3 8B     & 0.55 & \textbf{0.19} & 0.46 & 0.52 & 0.35 \\
Qwen3 14B    & 0.55 & 0.12 & 0.25 & 0.46 & 0.12 \\
\bottomrule
\end{tabular}

\caption{LHE Accuracy and Causality by model and domain, averaged across hierarchical levels on child--parent examples shared by all four models after LM-based word-prediction filtering.}
\label{tab:acc_causality_models}
\end{table}

\definecolor{greenface}{HTML}{ECFDF5}
\definecolor{greenedge}{HTML}{A7F3D0}
\definecolor{greentext}{HTML}{065F46}
\definecolor{redface}{HTML}{FEF2F2}
\definecolor{rededge}{HTML}{FECACA}
\definecolor{redtext}{HTML}{991B1B}
\definecolor{subtletext}{HTML}{374151}
\definecolor{purpletext}{HTML}{7C3AED}
\definecolor{orangetext}{HTML}{D97706}

\newcommand{\cmark}{\ding{51}}
\newcommand{\xmark}{\ding{55}}

\newcommand{\ok}[1]{%
  \begingroup\setlength{\fboxsep}{1.2pt}%
  \fcolorbox{greenedge}{greenface}{\textcolor{greentext}{#1}}%
  \endgroup%
}
\newcommand{\bad}[1]{%
  \begingroup\setlength{\fboxsep}{1.2pt}%
  \fcolorbox{rededge}{redface}{\textcolor{redtext}{#1}}%
  \endgroup%
}

\newcommand{\patAllCorrect}{\textcolor{greentext}{All correct}}
\newcommand{\patBottomXTopC}{\textcolor{purpletext}{Bottom \xmark, Top \cmark}}
\newcommand{\patBottomCTopX}{\textcolor{orangetext}{Bottom \cmark, Top \xmark}}
\newcommand{\patAllWrong}{\textcolor{redtext}{All \xmark}}

\begin{table}[t]
\centering
\small 
\begin{tabularx}{\columnwidth}{>{\raggedright\arraybackslash}X}
\toprule

\textbf{Cairo (Locations)}\hfill\textbf{\patAllCorrect}\par
Pred:\par
Cairo $\subset$ \ok{Egypt \cmark} $\subset$ \ok{Northern Africa \cmark} $\subset$ \ok{Africa \cmark}\par
Gold: Cairo $\subset$ \textcolor{subtletext}{Egypt} $\subset$ \textcolor{subtletext}{Northern Africa} $\subset$ \textcolor{subtletext}{Africa}
\\ \midrule

\textbf{Lionel Messi (Persons)}\hfill\textbf{\patAllCorrect}\par
Pred: Lionel Messi $\subset$ \ok{Soccer player \cmark} $\subset$ \ok{Athlete \cmark}\par
Gold: Lionel Messi $\subset$ \textcolor{subtletext}{Soccer player} $\subset$ \textcolor{subtletext}{Athlete}

\\ \midrule

\textbf{Leping (Locations)}\hfill\textbf{\patBottomXTopC}\par
Pred: Leping $\subset$ \bad{India \xmark} $\not\subset$ \ok{Eastern Asia \cmark} $\subset$ \ok{Asia \cmark}\par
Gold: Leping $\subset$ \textcolor{subtletext}{China} $\subset$ \textcolor{subtletext}{Eastern Asia} $\subset$ \textcolor{subtletext}{Asia}
\\ \midrule

\textbf{Dolphin (Organisms)}\hfill\textbf{\patBottomCTopX}\par
Pred: Dolphin $\subset$ \ok{Cetacea \cmark} $\not\subset$ \bad{Fish \xmark}\par
Gold: Dolphin $\subset$ \textcolor{subtletext}{Cetacea} $\subset$ \textcolor{subtletext}{Mammal}

\\ \midrule

\textbf{The New York Times Company (Organizations)}\par
\hfill\textbf{\patBottomXTopC}\par
Pred: The New York Times Company $\subset$ \bad{Consumer goods company \xmark} $\subset$ \ok{Company \cmark}\par
Gold: The New York Times Company $\subset$ \textcolor{subtletext}{Media company} $\subset$ \textcolor{subtletext}{Company}

\\
\bottomrule
\end{tabularx}
\caption{Accuracy examples for Llama 3.1 8B. Colors and \cmark/\xmark{}
mark step-wise correctness. Additional examples appear in
Tbl.~\ref{tab:accuracy_cases_appendix}.}
  \label{tab:accuracy_cases}
\end{table}

\begin{table}[t]
  \centering
  \small
  \begin{tabular}{cccc}
    \toprule
    \makecell{Child\\(outcome)} & Before & Target & After \\
    \midrule
    \makecell{Hank Aaron\\{\color{successlabel}\textbf{Success}}}
      & \makecell{Baseball\\player}
      & \makecell{Soccer\\player}
      & \makecell{Soccer\\player} \\
    \cmidrule(lr){1-4}
    \makecell{Silvassa\\{\color{successlabel}\textbf{Success}}}
      & India & Japan & Japan \\
    \midrule
    \makecell{LeBron James\\{\color{faillabel}\textbf{Failure}}}
      & \makecell{Basketball\\player}
      & \makecell{Tennis\\player}
      & \makecell{Basketball\\player} \\
    \bottomrule
  \end{tabular}
    \caption{Causality examples for Llama 3.1 8B. Before and After are
    the predicted parents before and after intervention; Target is the
    intended parent.}
  \label{tab:causality_cases}
\end{table}

\section{RQ1: Are Hierarchical Relations Linearly Encoded and Used?}
\label{sec:result}

We first show that LHE recovers features that predict parent relations and affect LM behavior.

\subsection{Comparison with Baselines}
We compare LHE with a linear support vector machine (SVM) and Input Averaging.
The SVM provides the stronger comparison by learning discriminative concept directions from labeled child representations.\footnote{Since SVMs cannot naturally predict unseen labels, we include a subset of the test data in training, which makes this a favorable setting for the SVM.}
Input Averaging is a simple baseline that uses the mean child representation for each parent as its concept vector.\footnote{We do not compare to \citet{geo_hierarchical_concepts}'s method because it is not applicable to intermediate layers.}
Tbl.~\ref{tab:acc_causality} compares LHE with the linear baselines on Llama 3.1 8B.
Across all four models and five domains, LHE outperforms both baselines in Accuracy in all 20 settings and achieves the highest Causality in 18 of the 20 settings (Tbl.~\ref{tab:acc_causality_all_models} in App.~\ref{sec:appendix_other_model_results}).%
\footnote{We also compared the target parent's rank among all answer choices before and after editing to see whether LHE moves it toward the top.
For Llama 3.1 8B, mean rank improved from 4.86 to 3.05, Hits@1 rose from 1.0\% to 31.1\%, and Hits@3 rose from 36.2\% to 67.8\%.
See App.~\ref{sec:appendix_target_parent_rank} for results on all four models.}
For example, on Llama 3.1 8B, LHE reaches 0.75 Accuracy and 0.56 Causality, compared with 0.55/0.35 for SVM and 0.40/0.16 for Input Averaging.
Thus, the child-to-parent transformation more reliably extracts features that support both linear decoding and causal steering than the linear baselines.
Representative successes and failures appear in Tables~\ref{tab:accuracy_cases} and~\ref{tab:causality_cases}.

\paragraph{Shuffled test parent labels.}
To test whether LHE's Accuracy could be explained by simply distinguishing lower-level concepts from higher-level ones without capturing child--parent correspondences, we used a shuffle control.
We fitted LHE on correct child--parent pairs and, at test time, randomly shuffled parent labels within the same domain and at the same depth.
Across the four models, Accuracy was 0.74--0.84 with the original parent labels and 0.007--0.018 with the shuffled parent labels (Tbl.~\ref{tab:relation_swap_control} in App.~\ref{sec:appendix_relation_swap_control}).
These results support the interpretation that LHE's Accuracy reflects not only the distinction between lower-level and higher-level concepts but also individual child--parent correspondences.

\paragraph{Interventions in Random Directions.}
To test whether intervention effects depend on direction, we replaced each LHE source--target displacement with a random direction of the same norm.
On Llama 3.1 8B, Causality was 0.0014 for random directions and 0.56 for LHE.
The other models showed the same qualitative trend (Tbl.~\ref{tab:random_direction_control} in App.~\ref{sec:appendix_random_direction_control}).
These results suggest that, in addition to intervention magnitude, the directions extracted by LHE are important for steering.

\subsection{Results across Models and Domains}
Tbl.~\ref{tab:acc_causality_models} summarizes the results.\footnote{Note that these results are computed on the intersection of the test instances that remain after word-prediction filtering for all LMs, and thus the test set differs from that used in Tbl.~\ref{tab:acc_causality}.}
We find that Accuracy ranges from approximately 0.5 to 0.9 across all models and domains.
Causality ranges from 0.35 to 0.7 in most settings, except for the research topic domain and Qwen3 14B.
Thus, hierarchical relations are linearly decodable across all four models, while their causal influence varies substantially.

\subsection{Linear Decodability and Causal Use Can Dissociate}
Llama 3.2 3B and Llama 3.1 8B score relatively highly on both metrics, whereas Qwen3 14B often shows reasonable Accuracy but low Causality.
In these settings, the parent relation can be recovered linearly from internal representations, but the directions identified by LHE only weakly control word predictions.
This dissociation between Accuracy and Causality reflects a well-known distinction in the probing literature between what is \emph{encoded} in a model's representations and what the model actually \emph{uses}: probing accuracy alone does not justify causal conclusions about model behavior \citep{elazar-etal-2021-amnesic}, and latent structure identified by probing can fail to explain downstream behavior \citep{agarwal-etal-2025-mechanisms}. Our Qwen3 14B results illustrate the same distinction in the context of hierarchical representations.
This variation motivates RQ2: where are the behaviorally relevant features located, and how are they organized?
Causality here tests only the directly edited relation; Sec.~\ref{sec:counterfactual} returns to RQ3 by testing effects on higher ancestors.

\section{RQ2: How Are Hierarchical Representations Organized?}
\label{sec:analysis}
\label{sec:organization}
We next examine \emph{how} that information is organized in the representation space.
We show that (i) hierarchical information is concentrated in a low-dimensional subspace, (ii) the causally relevant subspace is largely domain-specific, and (iii) despite this domain dependence, hierarchical representations share a similar structure across domains.
These analyses distinguish the \emph{location} of hierarchical representations from their \emph{organization}: different domains use different subspaces but organize hierarchical information similarly.

\subsection{Effective Dimensionality}
\begin{figure}[t]
\centering
\includegraphics[width=\linewidth]{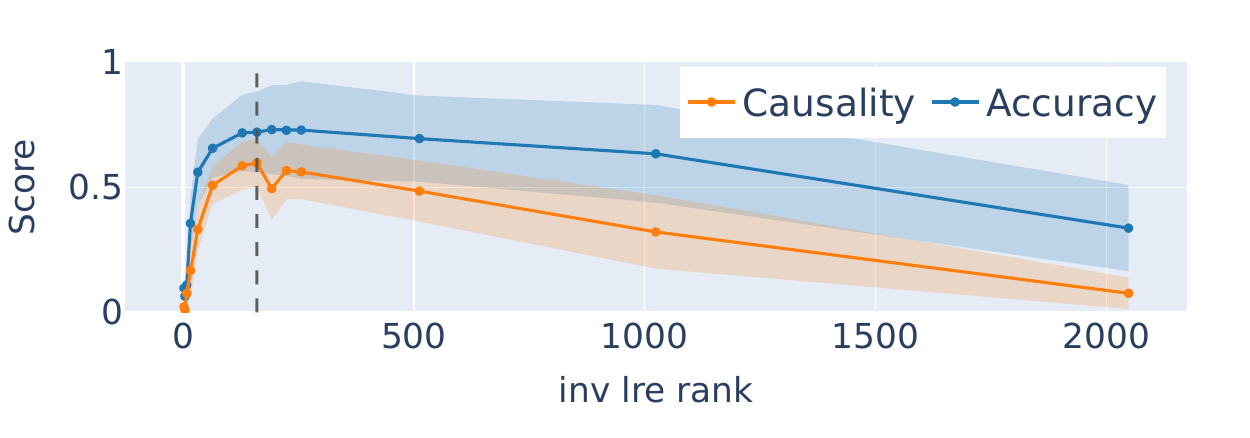}
\caption{
Post-hoc sweep of the truncation rank of $\mathbf{W}_r^{\dagger}$ for Llama 3.1 8B on the held-out test subtrees.
Scores are averaged across domains. The vertical dashed line marks the rank with the highest mean of test Accuracy and Causality.
App.~\ref{sec:appendix_sweep} shows similar trends across models.
}
\label{fig:sweep_rank}
\end{figure}

We ask how many dimensions of the learned mapping are useful for hierarchy decoding and intervention.
In a post-hoc sensitivity analysis on the held-out test subtrees, we vary the truncation rank $k$ of the pseudo-inverse $\mathbf{W}_r^{\dagger}$.
Rank $k$ retains the top $k$ singular components, yielding a $k$-dimensional intervention subspace.
As shown in Fig.~\ref{fig:sweep_rank}, both Accuracy and Causality peak at ranks of approximately 150--250.
This pattern suggests that relatively low ranks capture the hierarchy information most useful for decoding and intervention.

\subsection{Domain Specificity}
\label{sec:domain_specificity}
We first ask at what granularity the child-to-parent map must be estimated.
We compare a single global map, one map per depth shared across domains, one map per domain shared across depths, and the default LHE configuration with one map per domain--depth pair (details in App.~\ref{sec:appendix_map_ablation}).
Tbl.~\ref{tab:map_granularity_main} presents these configurations from the coarsest to the finest granularity.
Even a single global map retains 0.66 Accuracy on Llama 3.1 8B, but using one map per domain recovers substantially more Causality and is the strongest shared-map configuration across all four models (Tbl.~\ref{tab:map_granularity_all_models}).
Default LHE is best or tied-best in Accuracy for all four models and in Causality for three.
The contrast suggests that some hierarchy cues are globally decodable, whereas behaviorally effective directions depend more strongly on domain than on depth.

\begin{table}[t]
\centering
\small
\begin{tabular}{lccc}
\toprule
LHE map granularity & \# maps & Acc. & Caus. \\
\midrule
Global & 1 & 0.66 & 0.36 \\
Per depth (across domains) & 3 & 0.69 & 0.38 \\
Per domain (across depths) & 5 & 0.72 & 0.51 \\
Domain--depth (default) & 12 & \textbf{0.75} & \textbf{0.56} \\
\bottomrule
\end{tabular}
\caption{Map-granularity ablation for LHE on Llama 3.1 8B, ordered by the number of maps. Scores are macro-averaged over the five domains. Results for all four models are reported in Tbl.~\ref{tab:map_granularity_all_models} in App.~\ref{sec:appendix_map_results}.}
\label{tab:map_granularity_main}
\end{table}

\begin{figure}[t]
\centering
\includegraphics[width=\linewidth]{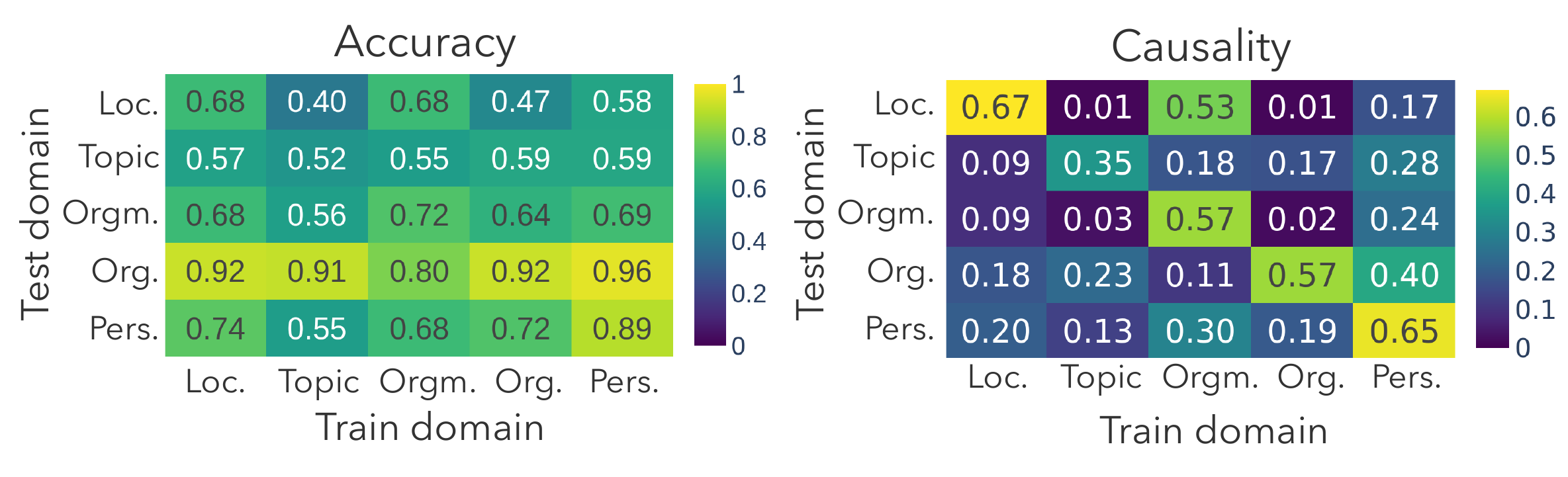}
\caption{Cross-domain evaluation of hierarchical-structure directions in Llama 3.1 8B. 
We train a linear transformation on a domain (x-axis) and evaluate it on test domains (y-axis).
Left: Accuracy. Right: Causality (intervention success). Causality drops sharply under domain shift, while Accuracy remains relatively stable.
Results for all models in Fig.~\ref{fig:domain_shift_all_model}.
}
\label{fig:domain_shift}
\end{figure}

We test this interpretation directly using default LHE by training a transformation on one domain and evaluating it on another.
Fig.~\ref{fig:domain_shift} shows that Accuracy remains relatively high for many cross-domain combinations, suggesting that some cues for decoding hierarchical labels transfer across domains.
In contrast, Causality is highest within the same domain (the diagonal entries of Fig.~\ref{fig:domain_shift}) and drops substantially under domain shift; for some domain pairs, interventions almost never succeed.
This mismatch between the patterns of Accuracy and Causality indicates that decoding hierarchical information from representations does not necessarily imply that the decoded features are also causally effective for prediction \citep{elazar-etal-2021-amnesic,agarwal-etal-2025-mechanisms}.
Instead, the intervenable features that affect prediction appear to be encoded differently across domains.

\begin{figure}[t]
\centering
\includegraphics[width=\linewidth]{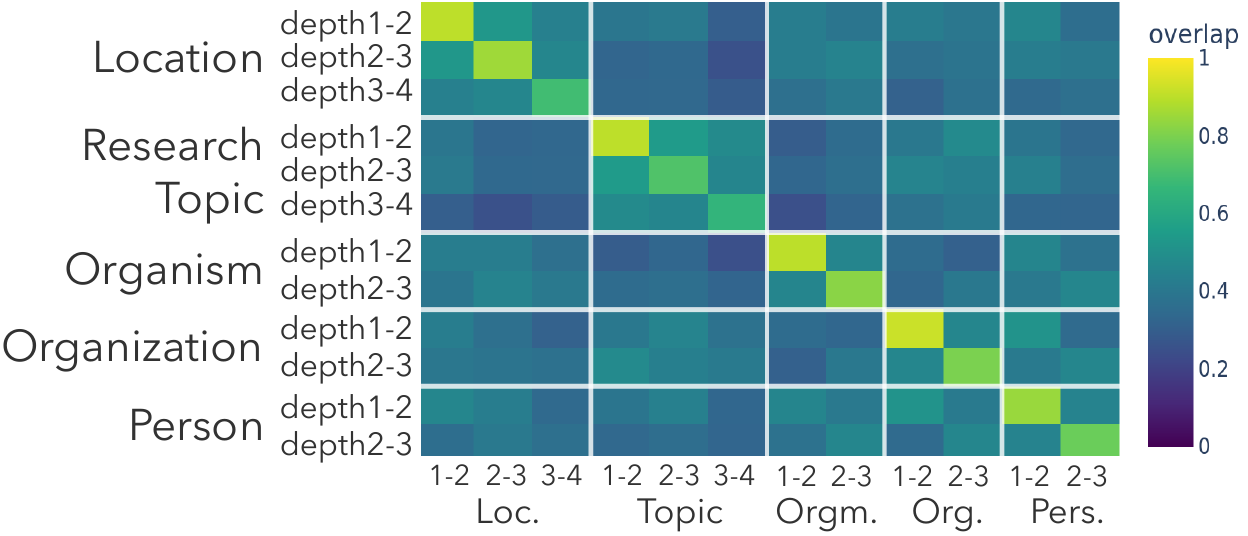}
\caption{Subspace similarity of subject representations for Llama 3.1 8B.
We decompose the learned linear transformation matrix $W$ via a low-rank SVD, $W \approx U\mathrm{diag}(s)V^\top$, and quantify the overlap between the subspaces spanned by the right singular vectors $V$ as
$\mathrm{overlap}(A,B)=\frac{1}{r}\|V_A^\top V_B\|_F^2$,
where $r$ denotes the rank and $A,B$ are the two conditions being compared.
Results for all models in Fig.~\ref{fig:subspace_sim_all_models}.
}
\label{fig:subspace_sim}
\end{figure}

For a more fine-grained analysis, we perform singular value decomposition of the learned linear transformation matrix $W$ and measure the similarity between the subject-side subspaces involved in causal interventions (Fig.~\ref{fig:subspace_sim}).
We find that subspace similarity consistently decreases across domains.
These results suggest that hierarchical concept representations that causally influence prediction are localized in domain-specific subspaces.

\begin{figure}[t]
\centering
\includegraphics[width=\linewidth]{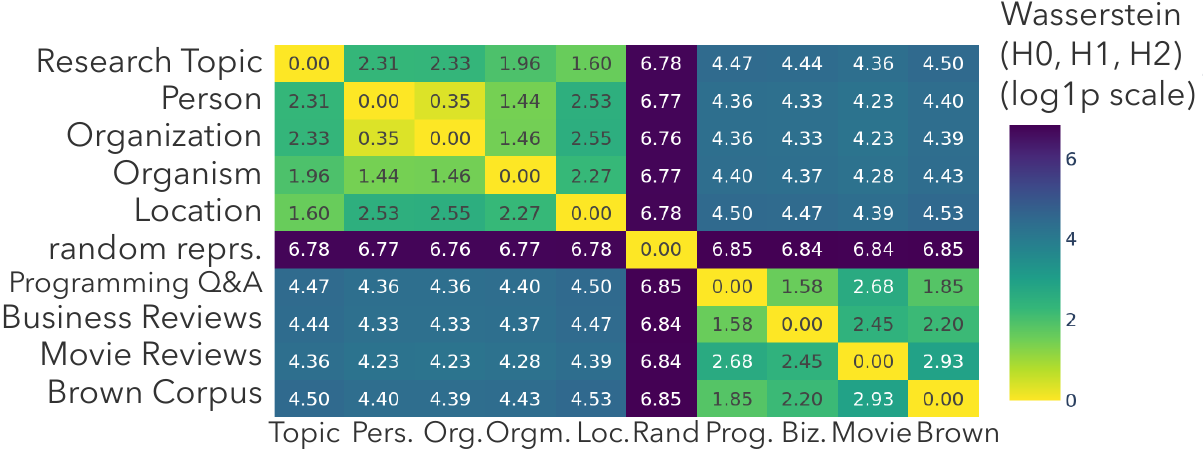}
\caption{Similarity matrix of LM representations based on persistent homology for Llama 3.1 8B. Each entry shows the Wasserstein distance, computed on persistence diagrams derived from concept vector point clouds, between a pair of domains or datasets. Smaller values indicate more similar representational structure.
Results for all models in Fig.~\ref{fig:tda_four_model}.}
\label{fig:tda}
\end{figure}

\subsection{Shared Organization across Domains}
The preceding results locate behaviorally relevant hierarchy information in different domain-specific subspaces.
We now ask whether those spaces nevertheless share an internal geometric organization.
As a qualitative analysis, we visualize the concept vectors obtained from the learned linear transformations using PCA.
Fig.~\ref{fig:pontie} provides an overview: its center panel shows domain-specific clustering, while its right panel shows that concept vectors in the Organism and Person domains form coherent arrangements reflecting their corresponding subtrees.
In the Organism example, higher-level concepts such as \emph{Animal}, \emph{Mammal}, and \emph{Bird} are separated from their lower-level descendants; the Person example exhibits an analogous organization under \emph{Worker}.
These examples suggest a two-level picture: concept vectors occupy domain-specific regions at the global scale, while exhibiting similar organization by subtree and hierarchical depth within each domain.
We observe the same qualitative pattern in the other domains; the corresponding PCA visualizations and domain-level examples are provided in App.~\ref{sec:appendix_pca}.

PCA depends on qualitative inspection, and ordinary representation-similarity measures require instance-level correspondences that do not exist between entities in different domains.
We therefore use persistent homology, a correspondence-free summary of point-cloud structure, and compare Wasserstein distances between the resulting persistence diagrams.
Intuitively, persistent homology treats each representation vector as a point and tracks changes in point-cloud structure as balls around the points gradually expand.
A persistence diagram records the scales at which groups of points linked by overlapping balls (connected components) merge and loops appear or disappear.
Smaller Wasserstein distances indicate more similar patterns of feature appearance and disappearance between the compared point clouds.
We contrast the five hierarchical domains with five baselines: random representations and four non-hierarchical text corpora (see App.~\ref{sec:appendix_tda} for details).
Fig.~\ref{fig:tda} reports the resulting distance matrix.
Distances are smaller between pairs of hierarchical domains than between a hierarchical domain and a baseline, indicating higher structural similarity.
For example, averaging over unique off-diagonal pairs, the mean distance is $1.88$ among hierarchical domains, compared with $4.87$ between a hierarchical domain and a baseline.
Overall, the subspaces differ across domains, but their internal point-cloud structure is similar.

\section{RQ3: Do Local Edits Produce Hierarchy-Consistent Effects?}
\label{sec:counterfactual}

\begin{figure}[t]
\centering
\includegraphics[width=\columnwidth]{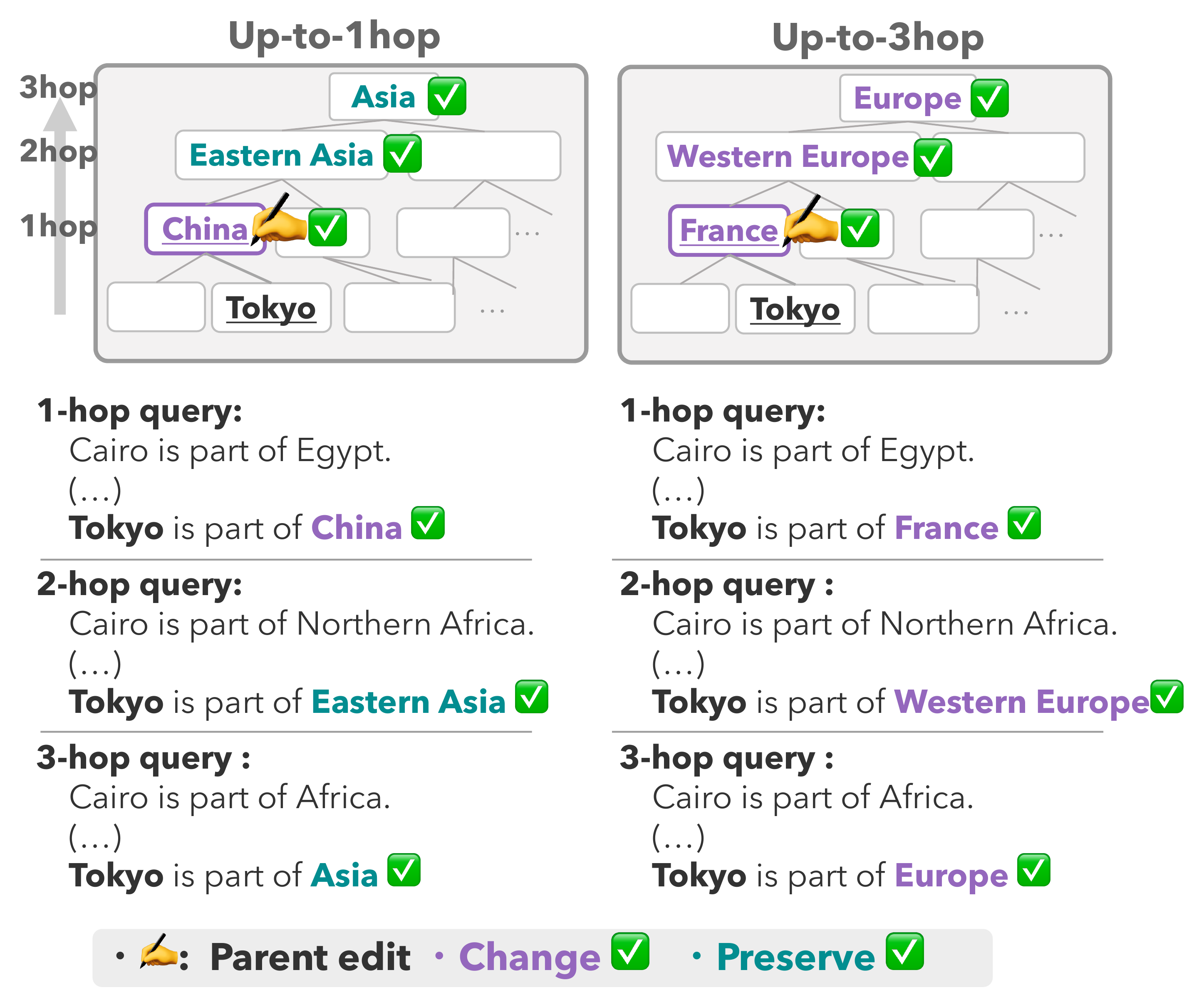}
\caption{
Illustration of counterfactual editing conditions. Under Up-to-1hop, only the direct parent changes while shared ancestors are preserved; under Up-to-3hop, higher ancestors also change.
}
\label{fig:counterfactual_conditions}
\end{figure}

Direct steering shows that individual parent directions affect predictions.
As the causal capstone of our study, we now ask whether those directions participate in a broader, internally coherent hierarchy: does editing only a direct-parent relation also affect predictions multiple hops above it, without separate interventions on the ancestor relations?
For example, if we edit \emph{Tokyo}'s direct parent from \emph{Japan} to \emph{France}, do its 2-hop and 3-hop predictions also change to \emph{Western Europe} and \emph{Europe}, following the counterfactual chain (Fig.~\ref{fig:counterfactual_conditions}, right)?
Conversely, if we edit its direct parent from \emph{Japan} to \emph{China}, do its 2-hop and 3-hop predictions remain \emph{Eastern Asia} and \emph{Asia}, which are shared by the original and counterfactual trees (Fig.~\ref{fig:counterfactual_conditions}, left)?
We thus test whether ancestor predictions follow the counterfactual hierarchy.

\paragraph{Setup.}
For each leaf item (e.g., \emph{Tokyo}), we intervene on its hidden representation to steer its direct-parent prediction from the original parent to a counterfactual parent.
Under this intervention, we separately query the ancestors of the same leaf at 1 hop, 2 hops, and 3 hops.
The queried ancestor distance (1, 2, or 3 hops) is specified by using ancestors at that distance as the gold labels in the few-shot demonstrations (illustrated in Fig.~\ref{fig:counterfactual_conditions}; details in App.~\ref{sec:appendix_counterfactual}).
For the \emph{Up-to-$k$hop condition}, we select a counterfactual parent such that the ancestor labels at distances 1 through $k$ are expected to \emph{Change} relative to the original path.
For example, under Up-to-3hop in Fig.~\ref{fig:counterfactual_conditions} (right), we intervene on \emph{Tokyo}'s hidden-state vector to steer its parent prediction toward \emph{France}.
We evaluate whether \emph{Change} occurs in predictions up to the 3-hop ancestor along the counterfactual path \emph{Tokyo} $\to$ \emph{France} $\to$ \emph{Western Europe} $\to$ \emph{Europe} when only this direct-parent relation is edited.
Change succeeds when the counterfactual label scores above the original label, whereas Preserve succeeds when the shared label remains top-ranked among the candidates.
Across four LMs and five domains, we compare LHE with SVM and Input Averaging using the hidden-state intervention and multi-token scoring from the Causality evaluation.

\paragraph{Hop-wise intervention results.}
\begin{table}[t]
\centering
\small
\begin{tabular}{llccc}
\toprule
Method & Condition & 1-hop & 2-hop & 3-hop \\
\midrule
\multirow{3}{*}{Input avg.} & Up-to-1hop & 0.29 (C) & \textbf{0.87} (P) & \textbf{0.87} (P) \\
 & Up-to-2hop & 0.12 (C) & 0.31 (C) & \textbf{0.86} (P) \\
 & Up-to-3hop & 0.12 (C) & 0.28 (C) & 0.35 (C) \\
\midrule
\multirow{3}{*}{SVM} & Up-to-1hop & 0.41 (C) & 0.81 (P) & 0.77 (P) \\
 & Up-to-2hop & 0.20 (C) & 0.40 (C) & 0.80 (P) \\
 & Up-to-3hop & 0.14 (C) & 0.33 (C) & 0.37 (C) \\
\midrule
\multirow{3}{*}{LHE} & Up-to-1hop & \textbf{0.42} (C) & 0.85 (P) & 0.85 (P) \\
 & Up-to-2hop & \textbf{0.32} (C) & \textbf{0.55} (C) & 0.85 (P) \\
 & Up-to-3hop & \textbf{0.30} (C) & \textbf{0.56} (C) & \textbf{0.63} (C) \\
\bottomrule
\end{tabular}
\caption{Post-intervention success rates by method, condition, and ancestor distance for Llama 3.1 8B on Locations. Each score is followed by its expected behavior (C: Change; P: Preserve). Bold marks the best method for each condition and ancestor distance. Complete results for all models and domains are reported in App.~\ref{sec:appendix_counterfactual}.}
\label{tab:counterfactual_hopwise_main}
\end{table}

Tbl.~\ref{tab:counterfactual_hopwise_main} compares the three methods by condition and ancestor distance for Llama 3.1 8B on Locations.
In this setting, Input Averaging achieves the highest success rate in every Preserve cell but the lowest rate in every Change cell.
This combination is consistent with limited intervention efficacy: predictions are retained, but often fail to move when a change is required.
LHE, by contrast, maintains high Preserve success (0.85 in all Preserve cells) while achieving the highest success rate in every Change cell.
In one successful case from Locations, the LHE intervention was constructed solely from the direct-parent edit for \emph{Tiberias}, from \emph{Israel} to \emph{Morocco}.
After the edit, \emph{Morocco} scored above \emph{Israel} when querying Tiberias's direct parent, \emph{Northern Africa} above \emph{Western Asia} when querying its 2-hop ancestor, and \emph{Africa} above \emph{Asia} when querying its 3-hop ancestor.
Thus, although only the direct-parent relation was edited, the predictions for Tiberias's 2-hop and 3-hop ancestors also moved to the labels specified by the counterfactual tree.
The macro-average over all available cells across domains, conditions, and ancestor distances shows the same Change advantage: LHE achieves the highest Change success rate on all four models (0.30--0.45, vs.~0.20--0.26 for Input Averaging and 0.16--0.30 for SVM; Tbl.~\ref{tab:counterfactual_method} in App.~\ref{sec:appendix_counterfactual}).
Overall, LHE is strongest when predictions must Change while retaining high Preserve performance, but success remains domain- and model-dependent.
The results show hierarchy-consistent effects on aggregate hop-wise predictions; they do not establish sample-level rewriting of a complete symbolic tree or step-by-step causal propagation through every intermediate ancestor.

\section{Conclusion}
Using \emph{Linear Hierarchical Encoding} (LHE), we studied multi-token concept hierarchies in the intermediate representations of four LMs across five domains.
Parent--child relations were linearly decodable, and the extracted directions causally steered predictions.
In a post-hoc test-set sweep for Llama 3.1 8B, decoding and steering peaked at ranks of approximately 150--250.
Causally effective features occupied largely domain-specific subspaces with similar internal geometry.
Direct-parent edits also produced partially hierarchy-consistent Change and Preserve effects on higher-ancestor predictions.
Together, these findings suggest that LMs encode behaviorally relevant hierarchies in domain-specific but similarly organized subspaces.

\newpage
\clearpage

\section*{Limitations}
Our study has three main limitations.
First, to control the problem setting, we restricted our analysis to hierarchies in which each child node belongs to a single parent node.
In practice, however, concept hierarchies can contain concepts that belong to multiple parent concepts.
Extending our framework to such multiple-parent relations is an important direction for future work.
Second, we analyze hierarchical representations under a linear approximation.
Our goal is to empirically investigate the extent to which hierarchical structure is linearly decodable from LM representations, and the extent to which linear interventions can change model output.
Our results show that hierarchical structure is linearly decodable and steerable to a nontrivial degree.
However, this does not imply that hierarchical representations in LMs are realized entirely by linear mechanisms.
While we focus on linear approximations as a first step, analyzing nonlinear mechanisms for representing hierarchical structure is an important avenue for future work.
Third, the effects of counterfactual interventions vary across domains and models, so we do not claim that hierarchical knowledge is uniformly editable.
In Sec.~\ref{sec:counterfactual}, we apply an intervention intended to change the direct parent and separately query the ancestors of the same leaf at each distance.
The results show that the intervention also affects predictions about higher ancestors.
However, the internal mechanisms underlying this apparent stepwise information propagation remain a topic for future research.

\section*{Acknowledgments}
We would like to thank the members of the Tohoku NLP Group for their insightful comments.
This work was supported by JSPS KAKENHI Grant Numbers JP25KJ0628, JP22H05106, and JP23K24910; JST BOOST Grant Number JPMJBY24F9; JST FOREST Grant Number JPMJFR2331; and AMED Grant Number JP24wm0625405.
This work was supported by ABCI 3.0, which is provided by AIST and AIST Solutions.

\bibliography{custom}

\appendix
\newpage
\clearpage

\section{Details of the Data}
\label{sec:appendix_data}
\begin{figure*}[t]
\centering
\includegraphics[width=0.8\linewidth]{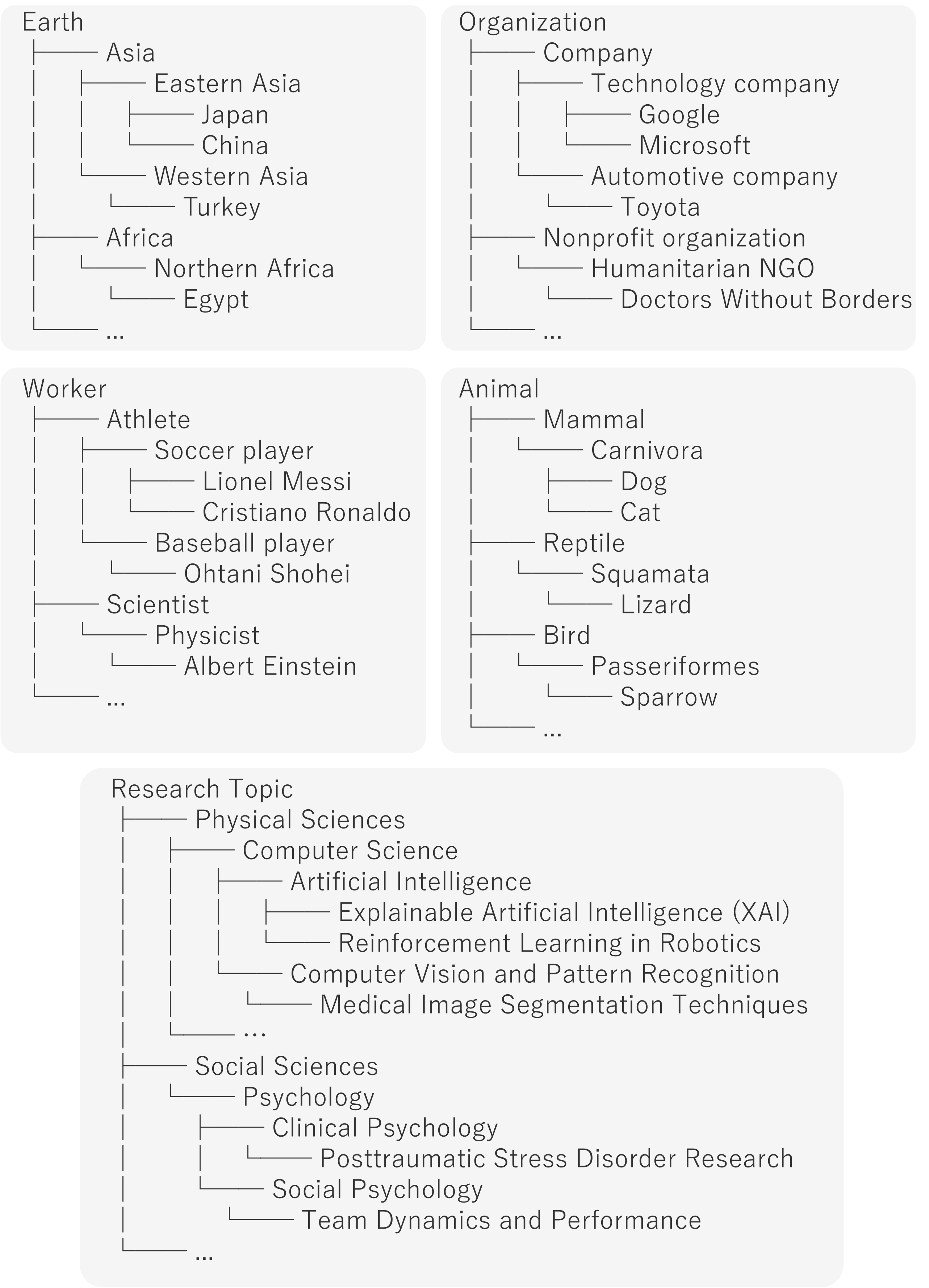}
\caption{Examples of data representing the hierarchical structure of concepts.}
\label{fig:data_example}
\end{figure*}

In this study, we use hierarchical datasets from five domains: Locations, Persons, Organizations, Organisms, and Research Topics.
For the location domain, we use the ISO 3166-1 country code dataset, which includes country classifications based on the United Nations geoscheme~\citep{lukes_iso3166_regional_codes_v10}, together with the geographic data used by \citet{repr_space}.
For the person and organization domains, we first generated candidate entities using GPT-5~\citep{gpt5} and then filtered them to retain only entries that exist in Wikidata~\citep{wikidata}, thereby ensuring that the entities correspond to real-world instances.
We chose to use GPT-5 for candidate generation because knowledge bases such as Wikidata sometimes contain category names introduced primarily for taxonomic convenience, including expressions that are not commonly used in the real world.
For the organism domain, we use the dataset of \citet{geo_hierarchical_concepts} and organize it according to the Linnaean taxonomic hierarchy.
For the research topic domain, we use publicly available data from OpenAlex~\citep{openalex}.
Concrete examples of the hierarchical data are shown in Fig.~\ref{fig:data_example}.

\paragraph{Data licenses.}
The ISO 3166-1 resource is distributed under the Creative Commons Attribution-ShareAlike 4.0 International license, the geographic data of \citet{repr_space} under the MIT License, and the OpenAlex and Wikidata data under the Creative Commons CC0 1.0 dedication.
The repository accompanying \citet{geo_hierarchical_concepts} does not specify a license; we use its organism data only for the research analyses described in this paper.

We then filtered the data to retain only those instances that each model answered correctly in a few-shot multiple-choice question answering setting.
The resulting number of filtered instances for each model is reported in Tbl.~\ref{tab:domain-depth-counts-all-models}.
The prompt used for this filtering step is shown in Fig.~\ref{fig:prompt}.
We used five few-shot examples during filtering.

\begin{table*}[t]
\centering
\small
\begin{tabular}{lrrrrr}
\toprule
Depth & Locations & Research Topic & Persons & Organizations & Organisms \\
\midrule
\multicolumn{6}{c}{\textbf{Model: Llama 3.2 3B}} \\
\midrule
0 (Root) & 1 & 1 & 1 & 1 & 1 \\
1 & 5 & 3 & 6 & 5 & 6 \\
2 & 21 & 12 & 31 & 14 & 40 \\
3 & 151 & 79 & 188 & 157 & 354 \\
4 & 20820 & 1302 & -- & -- & -- \\
\midrule
\multicolumn{6}{c}{\textbf{Model: Llama 3.1 8B}} \\
\midrule
0 (Root) & 1 & 1 & 1 & 1 & 1 \\
1 & 5 & 4 & 6 & 6 & 6 \\
2 & 22 & 19 & 33 & 18 & 48 \\
3 & 168 & 140 & 202 & 196 & 389 \\
4 & 25037 & 1959 & -- & -- & -- \\
\midrule
\multicolumn{6}{c}{\textbf{Model: Qwen3 8B}} \\
\midrule
0 (Root) & 1 & 1 & 1 & 1 & 1 \\
1 & 5 & 4 & 6 & 6 & 6 \\
2 & 22 & 17 & 33 & 18 & 48 \\
3 & 168 & 132 & 202 & 196 & 387 \\
4 & 21891 & 1800 & -- & -- & -- \\
\midrule
\multicolumn{6}{c}{\textbf{Model: Qwen3 14B}} \\
\midrule
0 (Root) & 1 & 1 & 1 & 1 & 1 \\
1 & 5 & 4 & 6 & 5 & 6 \\
2 & 22 & 19 & 34 & 16 & 46 \\
3 & 169 & 143 & 205 & 182 & 385 \\
4 & 22416 & 1913 & -- & -- & -- \\
\bottomrule
\end{tabular}
\caption{Number of LM-filtered examples by hierarchical depth across domains and models, where parent-child relations are retained only when the model correctly predicts the parent category.}
\label{tab:domain-depth-counts-all-models}
\end{table*}

\section{Details of the Training Setup}
\label{sec:appendix_train}
Our method builds on the LRC framework of \citet{lrc}.
In the main experiments, we estimated each linear transformation using eight training examples, matching the sample count in the LRE experiments of \citet{lre}.
We sampled the training examples separately for each of five random seeds and report results averaged across seeds.
In the main experiments, we used a few-shot prompt with five in-context examples for each sample.
We used the same prompt format at evaluation time.
Each job used a single NVIDIA H200 SXM GPU with 141 GB of memory.
Most model--domain--depth jobs, each covering five random seeds and evaluation, took approximately 1--11 minutes.
The largest depth-4 settings took up to approximately 110 minutes because they involved substantially larger evaluation sets.
Based on the available accounting records, the total computational budget, including exploratory and unsuccessful runs, was approximately 900 GPU-hours.

\paragraph{Model licenses and software.}
The Llama 3.1 and Llama 3.2 weights were used under the corresponding Llama Community Licenses, and the Qwen3 weights under the Apache License 2.0.
Experiments used Python 3.10.19, PyTorch 2.9.1, Transformers 4.57.3, scikit-learn 1.4.1.post1, and \texttt{linear-relational} 0.6.2.
The persistent-homology analyses used Ripser 0.6.14 and Persim 0.3.8, and the Gromov--Wasserstein analyses used POT 0.9.6.post1.

\section{Map-Granularity Ablation Details}
\label{sec:appendix_map_ablation}
The ablation covers 12 domain--depth pairs. Persons, Organisms, and Organizations each contain the transitions depth $2\to1$ and $3\to2$, while Locations and Research Topics additionally contain depth $4\to3$. Default LHE therefore has 12 maps; the per-domain, per-depth, and global configurations have five, three, and one, respectively.

We construct the pooled training data with sampling seed 42. For every domain--depth pair, we sample without replacement up to ten triples from its training split (or all triples if fewer than ten are available), retain their original domain and depth metadata, and merge the prompt templates of the source relations. We then pool the sampled triples according to the sharing scheme. This produces the estimation pools summarized in Tbl.~\ref{tab:map_ablation_pools}.

\begin{table}[t]
\centering
\small
\setlength{\tabcolsep}{4pt}
\begin{tabular}{llr}
\toprule
LHE map granularity & Pool & \# triples \\
\midrule
Global & All pairs & 117 \\
\midrule
\multirow{3}{*}{Per depth}
 & Depth $2\to1$ & 47 \\
 & Depth $3\to2$ & 50 \\
 & Depth $4\to3$ & 20 \\
\midrule
\multirow{5}{*}{Per domain}
 & Locations & 30 \\
 & Research Topics & 28 \\
 & Persons & 20 \\
 & Organisms & 20 \\
 & Organizations & 19 \\
\bottomrule
\end{tabular}
\caption{Training pools for the map-granularity ablation.}
\label{tab:map_ablation_pools}
\end{table}

Each pool defines a candidate training set. A shared map is estimated from examples sampled from that pool using the same average-Jacobian LRE procedure as default LHE, with training seeds 42--46. For Llama 3.2 3B, Llama 3.1 8B, and Qwen3 8B, the estimator uses eight examples per seed. The Qwen3 14B shared-map runs use four examples and two in-context examples because the standard setting exceeded GPU memory. After map estimation, its pseudo-inverse is applied separately to the parent representations from each constituent domain--depth pair; hence only the map is shared, not the relation- and parent-specific concept vectors.

Within each model, all six evaluated configurations use the same model-filtered test instances, taxonomy, and few-shot source at evaluation. The shared maps are estimated from the pooled, unfiltered training source described above, whereas default LHE is estimated from its model-filtered training source. This mismatch places the shared-map configurations at a disadvantage relative to default LHE. We therefore compare mapping granularities within each model and do not interpret absolute score differences between models from these runs.

\begin{table}[t]
\centering
\small
\begin{tabular}{lccc}
\toprule
Models                  & Hidden dim. & \#Layer & \#Head \\ 
\cmidrule(lr){1-1}  \cmidrule(lr){2-2} \cmidrule(lr){3-3} \cmidrule(lr){4-4}
Llama 3.2 3B          & 3072  & 28 & 24 \\
Llama 3.1 8B     & 4096 & 32 & 32 \\
Qwen3 8B    & 4096 & 36 & 32 \\
Qwen3 14B      & 5120 & 40 & 40 \\
\bottomrule
\end{tabular}
\caption{Hyperparameters of each model's architecture.}
\label{tab:models_detail}
\end{table}

\begin{figure*}[t]
\centering
\includegraphics[width=0.8\linewidth]{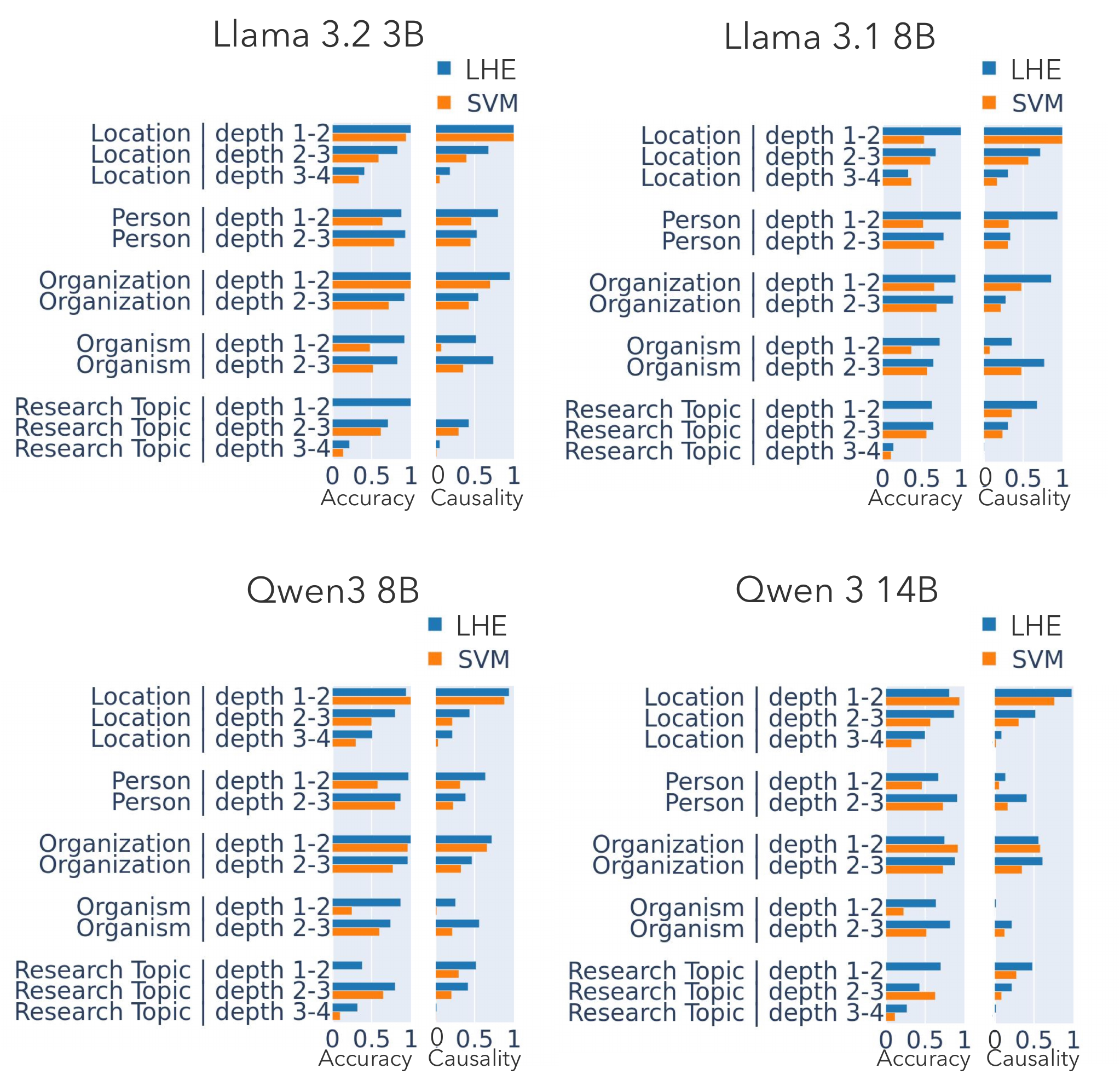}
\caption{Accuracy and Causality at each hierarchy depth for all models. The Accuracy reported in this figure is chance-corrected to account for the depth-specific chance rate and to enable fair comparison across depths.}
\label{fig:depth_summary}
\end{figure*}

\begin{figure*}[t]
\centering
\includegraphics[width=0.7\linewidth]{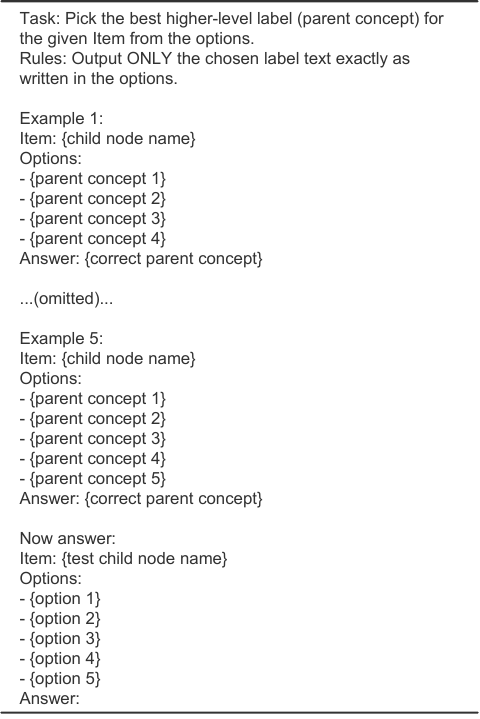}
\caption{Prompt format used for both data filtering and the main experiments.}
\label{fig:prompt}
\end{figure*}

\begin{figure*}[t]
\centering
\includegraphics[width=0.7\linewidth]{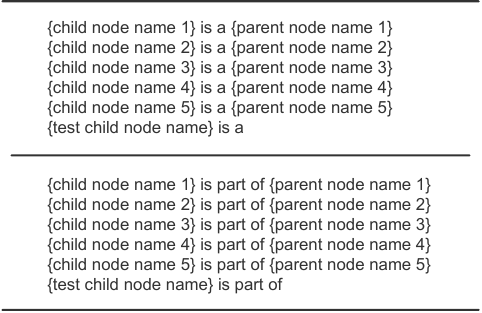}
\caption{Few-shot prompts in the template-based format. The is-a template is used for the Person, Organization, Organism, and Research Topic domains, whereas the part-of template is used for the Location domain. Results obtained with these prompts are shown in Tbl.~\ref{tab:acc_causality_all_models_templates_fewshot}.}
\label{fig:prompt_templates_fewshot}
\end{figure*}

\section{Hyperparameters for Each Model Used in the Experiments}
\label{sec:appendix_hyperparameter}
We split the training partition described in Sec.~\ref{sec:Experiments_setup} into train and dev.
For each model, we ran a grid search, fitting LHE on train and selecting the configuration with the highest arithmetic mean of Accuracy and Causality on dev.
We then refit LHE on the combined train and dev splits with the selected configuration and evaluated it on the held-out test subtrees.
Tbl.~\ref{tab:lrc_hyperparameter} lists the selected truncation rank, subject layer, object layer, and intervention coefficient $\beta$ for each model.

Models may differ in their sensitivity to interventions, so using a common intervention coefficient $\beta$ does not necessarily make cross-model comparisons fairer.
A fixed coefficient may favor models for which that value is near-optimal.
We therefore select $\beta$ separately for each model using the same dev-based selection criterion described above.
The reported scores thus reflect performance under model-specific, dev-selected configurations rather than at a common intervention coefficient.

\begin{table}[t]
  \centering
  \small
  \setlength{\tabcolsep}{3pt}
  \begin{tabular}{lcccc}
    \toprule
    Model & Subj. layer & Obj. layer & Rank & $\beta$ \\
    \midrule
    Llama 3.2 3B & 12 & 15 & 128 & 0.100 \\
    Llama 3.1 8B & 12 & 21 & 160 & 0.085 \\
    Qwen3 8B      & 20 & 27 & 192 & 0.100 \\
    Qwen3 14B     & 25 & 28 & 256 & 0.125 \\
    \bottomrule
  \end{tabular}
  \caption{Hyperparameters used for LRC training and inference for each model. Here, $\beta$ is the intervention coefficient used for Causality.}
  \label{tab:lrc_hyperparameter}
\end{table}

\section{Results for All Models}
\label{sec:appendix_other_model_results}
\subsection{Map-Granularity Ablation Across Models}
\label{sec:appendix_map_results}
Tbl.~\ref{tab:map_granularity_all_models} reports the macro-averaged ablation results for all four models, and Tbl.~\ref{tab:map_granularity_by_domain} gives the domain-level breakdown for the three shared-map configurations. The corresponding domain-level Input Averaging, SVM, and default LHE results, including standard deviations, are reported in Tbl.~\ref{tab:acc_causality_all_models}. Default LHE has the best or tied-best Accuracy in every model. Moreover, Causality consistently increases from global to per-depth to per-domain maps, while splitting the per-domain configuration further by depth changes Accuracy by at most 0.03. These results show that LHE retains substantial Accuracy with one global map, while sharing maps across domains loses behaviorally relevant information.

The pattern is not strictly monotonic for every metric. In particular, Qwen3 14B obtains lower Accuracy with per-depth maps than with a global map (0.63 vs.\ 0.68), primarily in Locations and Research Topics. Furthermore, default LHE Causality is 0.03 below the per-domain configuration for Qwen3 8B and tied with it for Qwen3 14B; we do not interpret these small differences as evidence that depth-specific maps are harmful. We therefore interpret the robust contrast as the cost of sharing maps \emph{across domains}; the results do not show that finer depth-specific partitioning always helps.

\begin{table*}[t]
\centering
\small
\setlength{\tabcolsep}{4pt}
\begin{tabular}{lcccccccc}
\toprule
& \multicolumn{2}{c}{Llama 3.2 3B} & \multicolumn{2}{c}{Llama 3.1 8B}
& \multicolumn{2}{c}{Qwen3 8B} & \multicolumn{2}{c}{Qwen3 14B} \\
\cmidrule(lr){2-3}\cmidrule(lr){4-5}\cmidrule(lr){6-7}\cmidrule(lr){8-9}
LHE map granularity & Acc. & Caus. & Acc. & Caus. & Acc. & Caus. & Acc. & Caus. \\
\midrule
Global                  & 0.78 & 0.53 & 0.66 & 0.36 & 0.75 & 0.48 & 0.68 & 0.31 \\
Per depth               & 0.78 & 0.53 & 0.69 & 0.38 & 0.80 & 0.49 & 0.63 & 0.31 \\
Per domain              & 0.82 & 0.55 & 0.72 & 0.51 & \textbf{0.81} & \textbf{0.50} & 0.71 & \textbf{0.35} \\
Per domain--depth pair (default) & \textbf{0.84} & \textbf{0.57} & \textbf{0.75} & \textbf{0.56} & \textbf{0.81} & 0.47 & \textbf{0.74} & \textbf{0.35} \\
\bottomrule
\end{tabular}
\caption{Map-granularity ablation for LHE across all models, ordered from the coarsest to the finest granularity. Per-depth maps are shared across domains, per-domain maps are shared across depths, and the per-domain--depth-pair row is default LHE. Accuracy and Causality are macro-averaged over the five domains. Bold indicates the best value for each model and metric. Rows within each model use identical test instances, taxonomies, and few-shot sources.}
\label{tab:map_granularity_all_models}
\end{table*}

\begin{table*}[t]
\centering
\small
\setlength{\tabcolsep}{4pt}
\begin{tabular}{lcccccccccc}
\toprule
& \multicolumn{5}{c}{Accuracy} & \multicolumn{5}{c}{Causality} \\
\cmidrule(lr){2-6}\cmidrule(lr){7-11}
LHE map granularity & Loc. & Topic & Pers. & Org. & Orgm. & Loc. & Topic & Pers. & Org. & Orgm. \\
\midrule
\multicolumn{11}{l}{\textbf{Llama 3.2 3B}} \\
Global          & 0.71 & 0.60 & 0.90 & 0.93 & 0.75 & 0.57 & 0.18 & 0.68 & 0.61 & 0.61 \\
Per depth       & 0.73 & 0.61 & 0.88 & 0.94 & 0.74 & 0.64 & 0.15 & 0.66 & 0.60 & 0.60 \\
Per domain      & 0.75 & 0.62 & 0.96 & 0.98 & 0.81 & 0.63 & 0.17 & 0.58 & 0.74 & 0.63 \\
\midrule
\multicolumn{11}{l}{\textbf{Llama 3.1 8B}} \\
Global          & 0.63 & 0.57 & 0.53 & 0.91 & 0.67 & 0.45 & 0.32 & 0.39 & 0.38 & 0.25 \\
Per depth       & 0.63 & 0.55 & 0.67 & 0.93 & 0.67 & 0.44 & 0.30 & 0.41 & 0.46 & 0.31 \\
Per domain      & 0.65 & 0.49 & 0.78 & 0.94 & 0.74 & 0.65 & 0.26 & 0.63 & 0.54 & 0.49 \\
\midrule
\multicolumn{11}{l}{\textbf{Qwen3 8B}} \\
Global          & 0.72 & 0.58 & 0.70 & 0.96 & 0.79 & 0.62 & 0.25 & 0.46 & 0.61 & 0.48 \\
Per depth       & 0.75 & 0.53 & 0.90 & 0.96 & 0.86 & 0.58 & 0.28 & 0.49 & 0.64 & 0.46 \\
Per domain      & 0.74 & 0.60 & 0.90 & 0.97 & 0.85 & 0.58 & 0.28 & 0.54 & 0.66 & 0.46 \\
\midrule
\multicolumn{11}{l}{\textbf{Qwen3 14B}} \\
Global          & 0.54 & 0.56 & 0.88 & 0.73 & 0.71 & 0.27 & 0.30 & 0.42 & 0.37 & 0.18 \\
Per depth       & 0.48 & 0.40 & 0.82 & 0.77 & 0.69 & 0.33 & 0.22 & 0.44 & 0.42 & 0.15 \\
Per domain      & 0.67 & 0.49 & 0.89 & 0.79 & 0.71 & 0.47 & 0.23 & 0.40 & 0.51 & 0.15 \\
\bottomrule
\end{tabular}
\caption{Domain-level results for the three shared-map granularities of LHE, ordered from the coarsest to the finest. Corresponding Input Averaging, SVM, and default LHE results are reported in Tbl.~\ref{tab:acc_causality_all_models}. Loc., Topic, Pers., Org., and Orgm. denote Locations, Research Topics, Persons, Organizations, and Organisms, respectively.}
\label{tab:map_granularity_by_domain}
\end{table*}

\subsection{LHE Results and Prompt Variants}
\begin{figure*}[t]
\centering
\includegraphics[width=\linewidth]{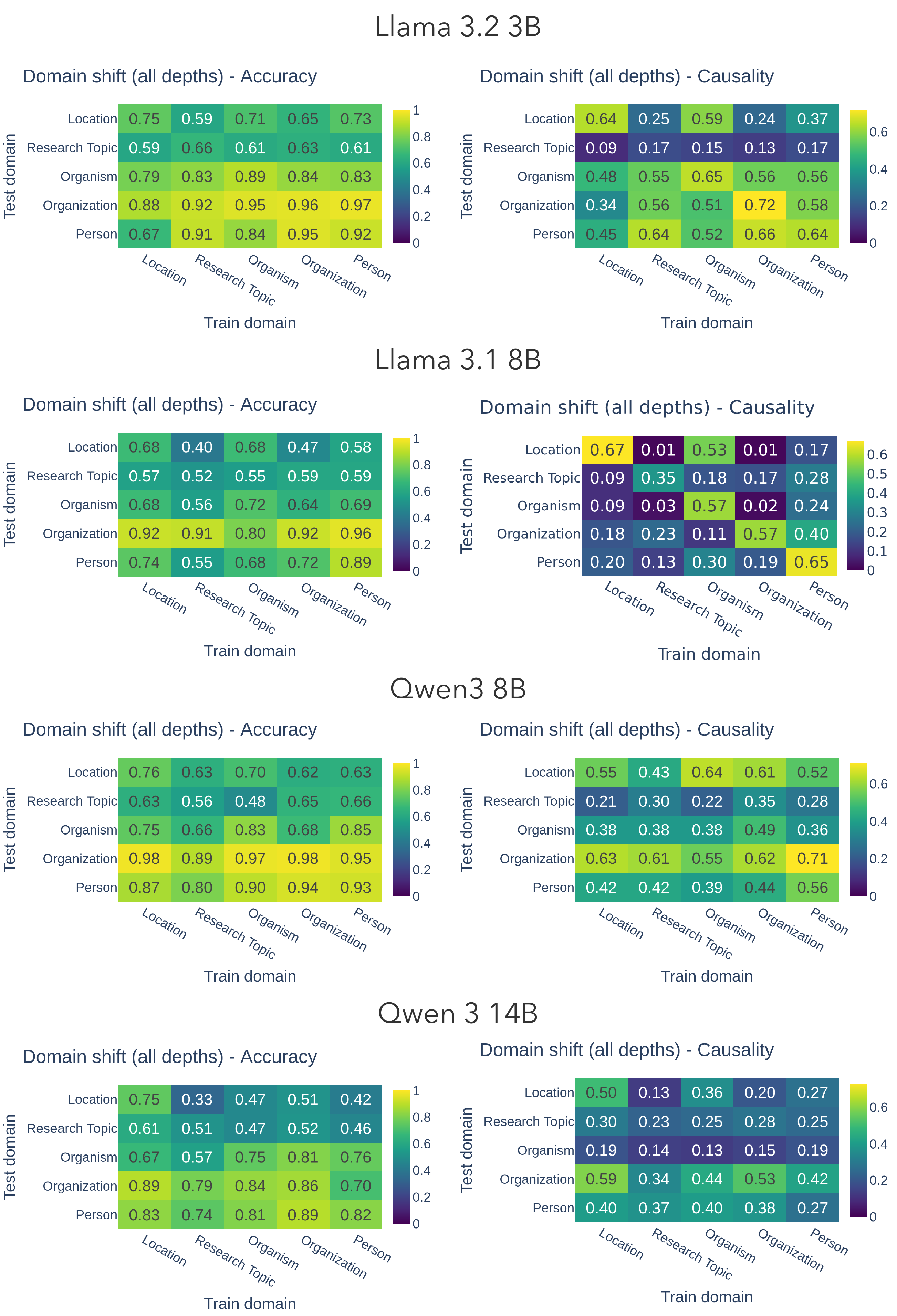}
\caption{Cross-domain evaluation of hierarchical-structure directions for all models.}
\label{fig:domain_shift_all_model}
\end{figure*}

\begin{figure*}[t]
\centering
\includegraphics[width=\linewidth]{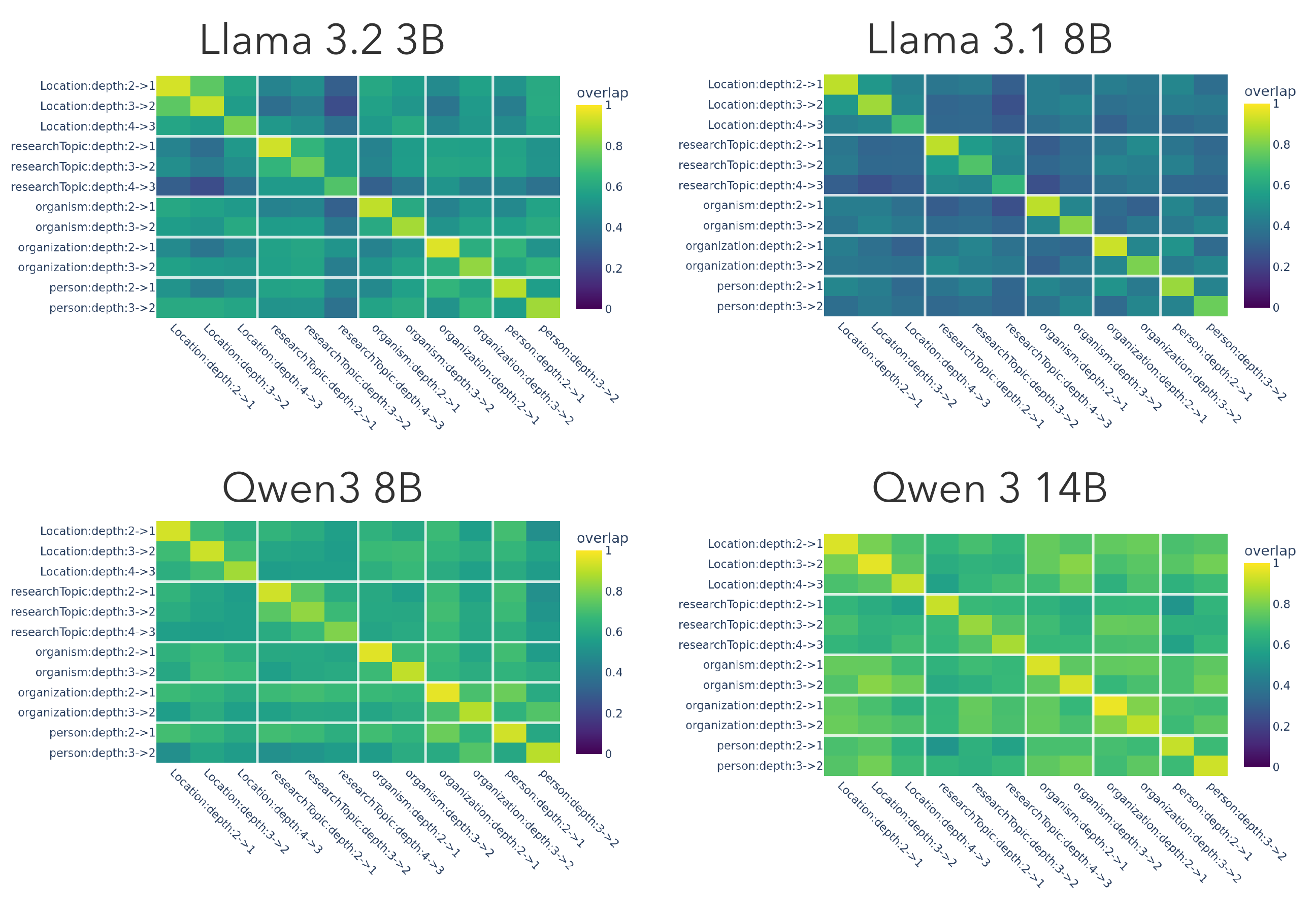}
\caption{Subspace similarity of subject representations for all models.
}
\label{fig:subspace_sim_all_models}
\end{figure*}

Tbl.~\ref{tab:acc_causality_all_models} reports the Accuracy and Causality results for all models.
In addition, Tbl.~\ref{tab:acc_causality_all_models_templates_fewshot} presents the results obtained using prompt variants with different template formats, shown in Fig.~\ref{fig:prompt_templates_fewshot}.
Fig.~\ref{fig:domain_shift_all_model} shows the results of the cross-domain evaluation for all models.
Fig.~\ref{fig:subspace_sim_all_models} shows the subspace similarity results for all models.

\begin{table*}[t]
\centering
\small 
\begin{tabular}{@{}p{\textwidth}@{}}
\toprule

\textbf{[Organism] panther}\hfill\textbf{\patAllCorrect}\par
Pred: panther $\subset$ \ok{Carnivora \cmark} $\subset$ \ok{Mammal \cmark}\par
Gold: panther $\subset$ \textcolor{subtletext}{Carnivora} $\subset$ \textcolor{subtletext}{Mammal}

\\ \midrule

\textbf{[Organization] Toyota}\hfill\textbf{\patAllCorrect}\par
Pred: Toyota $\subset$ \ok{Automotive company \cmark} $\subset$ \ok{Company \cmark}\par
Gold: Toyota $\subset$ \textcolor{subtletext}{Automotive company} $\subset$ \textcolor{subtletext}{Company}

\\ \midrule

\textbf{[Research Topic] Information Retrieval and Search Behavior}\hfill\textbf{\patBottomCTopX}\par
Pred: Information Retrieval and Search Behavior\par
\hspace*{1.5em}$\subset$ \ok{Information Systems \cmark} $\subset$ \ok{Computer Science \cmark} $\not\subset$ \bad{Social Sciences \xmark}\par
Gold: Information Retrieval and Search Behavior\par
\hspace*{1.5em}$\subset$ \textcolor{subtletext}{Information Systems} $\subset$ \textcolor{subtletext}{Computer Science} $\subset$ \textcolor{subtletext}{Physical Sciences}

\\
\bottomrule
\end{tabular}
\caption{Qualitative examples for accuracy using Llama 3.1 8B.
Each row shows an entity and its predicted parent chain (\textbf{Pred}) with step-wise correctness (\cmark/\xmark) against the gold chain (\textbf{Gold}).
Correct and incorrect predicted parents are highlighted in green and red, respectively, and $\subset$ / $\not\subset$ indicate inclusion vs.\ non-inclusion relations between consecutive depths.
We group cases by whether errors occur earlier vs.\ later along the evaluated depth sequence; this table shows representative instances of \textit{All correct}, \textit{Bottom \xmark, Top \cmark}, and \textit{Bottom \cmark, Top \xmark}.}
  \label{tab:accuracy_cases_appendix}
\end{table*}

\begin{table*}[t]
\centering
\small

\textbf{Llama 3.2 3B}\\
\textbf{Accuracy}\\
\begin{tabular}{lccccc}
\toprule
Method & Locations & Research Topic & Persons & Organizations & Organisms \\
\midrule
Input averaging & 0.47 $\pm$ 0.04 & 0.36 $\pm$ 0.15 & 0.60 $\pm$ 0.19 & 0.63 $\pm$ 0.21 & 0.48 $\pm$ 0.18 \\
SVM & 0.64 $\pm$ 0.03 & 0.37 $\pm$ 0.14 & 0.75 $\pm$ 0.14 & 0.87 $\pm$ 0.03 & 0.56 $\pm$ 0.20 \\
LHE & \textbf{0.75 $\pm$ 0.03} & \textbf{0.66 $\pm$ 0.04} & \textbf{0.92 $\pm$ 0.10} & \textbf{0.96 $\pm$ 0.02} & \textbf{0.89 $\pm$ 0.04} \\
\bottomrule
\end{tabular}

\vspace{4pt}

\textbf{Causality}\\
\begin{tabular}{lccccc}
\toprule
Method & Locations & Research Topic & Persons & Organizations & Organisms \\
\midrule
Input averaging & 0.43 $\pm$ 0.06 & 0.07 $\pm$ 0.02 & 0.24 $\pm$ 0.14 & 0.10 $\pm$ 0.01 & 0.16 $\pm$ 0.02 \\
SVM & 0.49 $\pm$ 0.01 & 0.11 $\pm$ 0.02 & 0.46 $\pm$ 0.16 & 0.56 $\pm$ 0.21 & 0.22 $\pm$ 0.04 \\
LHE & \textbf{0.62 $\pm$ 0.04} & \textbf{0.17 $\pm$ 0.06} & \textbf{0.67 $\pm$ 0.14} & \textbf{0.75 $\pm$ 0.02} & \textbf{0.63 $\pm$ 0.05} \\
\bottomrule
\end{tabular}

\vspace{10pt}

\textbf{Llama 3.1 8B}\\
\textbf{Accuracy}\\
\begin{tabular}{lccccc}
\toprule
Method & Locations & Research Topic & Persons & Organizations & Organisms \\
\midrule
Input averaging & 0.40 $\pm$ 0.11 & 0.27 $\pm$ 0.10 & 0.53 $\pm$ 0.26 & 0.36 $\pm$ 0.14 & 0.42 $\pm$ 0.15 \\
SVM & 0.54 $\pm$ 0.07 & 0.32 $\pm$ 0.09 & 0.64 $\pm$ 0.18 & 0.71 $\pm$ 0.06 & 0.54 $\pm$ 0.13 \\
LHE & \textbf{0.68 $\pm$ 0.03} & \textbf{0.52 $\pm$ 0.06} & \textbf{0.89 $\pm$ 0.03} & \textbf{0.93 $\pm$ 0.05} & \textbf{0.72 $\pm$ 0.07} \\
\bottomrule
\end{tabular}

\vspace{4pt}

\textbf{Causality}\\
\begin{tabular}{lccccc}
\toprule
Method & Locations & Research Topic & Persons & Organizations & Organisms \\
\midrule
Input averaging & 0.42 $\pm$ 0.07 & 0.05 $\pm$ 0.01 & 0.10 $\pm$ 0.09 & 0.07 $\pm$ 0.07 & 0.18 $\pm$ 0.07 \\
SVM & 0.58 $\pm$ 0.02 & 0.20 $\pm$ 0.06 & 0.32 $\pm$ 0.06 & 0.35 $\pm$ 0.11 & 0.28 $\pm$ 0.09 \\
LHE & \textbf{0.68 $\pm$ 0.03} & \textbf{0.33 $\pm$ 0.11} & \textbf{0.64 $\pm$ 0.05} & \textbf{0.57 $\pm$ 0.06} & \textbf{0.57 $\pm$ 0.08} \\
\bottomrule
\end{tabular}

\vspace{10pt}

\textbf{Qwen3 8B}\\
\textbf{Accuracy}\\
\begin{tabular}{lccccc}
\toprule
Method & Locations & Research Topic & Persons & Organizations & Organisms \\
\midrule
Input averaging & 0.55 $\pm$ 0.01 & 0.28 $\pm$ 0.06 & 0.69 $\pm$ 0.11 & 0.81 $\pm$ 0.07 & 0.44 $\pm$ 0.18 \\
SVM & 0.61 $\pm$ 0.02 & 0.31 $\pm$ 0.04 & 0.73 $\pm$ 0.07 & 0.88 $\pm$ 0.03 & 0.50 $\pm$ 0.14 \\
LHE & \textbf{0.76 $\pm$ 0.03} & \textbf{0.56 $\pm$ 0.08} & \textbf{0.93 $\pm$ 0.05} & \textbf{0.98 $\pm$ 0.01} & \textbf{0.82 $\pm$ 0.05} \\
\bottomrule
\end{tabular}

\vspace{4pt}

\textbf{Causality}\\
\begin{tabular}{lccccc}
\toprule
Method & Locations & Research Topic & Persons & Organizations & Organisms \\
\midrule
Input averaging & 0.41 $\pm$ 0.07 & 0.12 $\pm$ 0.06 & 0.15 $\pm$ 0.08 & 0.33 $\pm$ 0.13 & 0.19 $\pm$ 0.05 \\
SVM & 0.38 $\pm$ 0.06 & 0.17 $\pm$ 0.05 & 0.28 $\pm$ 0.08 & 0.49 $\pm$ 0.12 & 0.12 $\pm$ 0.02 \\
LHE & \textbf{0.53 $\pm$ 0.04} & \textbf{0.32 $\pm$ 0.13} & \textbf{0.51 $\pm$ 0.11} & \textbf{0.60 $\pm$ 0.05} & \textbf{0.41 $\pm$ 0.11} \\
\bottomrule
\end{tabular}

\vspace{10pt}

\textbf{Qwen3 14B}\\
\textbf{Accuracy}\\
\begin{tabular}{lccccc}
\toprule
Method & Locations & Research Topic & Persons & Organizations & Organisms \\
\midrule
Input averaging & 0.47 $\pm$ 0.05 & 0.29 $\pm$ 0.09 & 0.58 $\pm$ 0.14 & 0.56 $\pm$ 0.22 & 0.42 $\pm$ 0.21 \\
SVM & 0.63 $\pm$ 0.04 & 0.32 $\pm$ 0.08 & 0.65 $\pm$ 0.10 & 0.84 $\pm$ 0.05 & 0.45 $\pm$ 0.18 \\
LHE & \textbf{0.74 $\pm$ 0.08} & \textbf{0.51 $\pm$ 0.06} & \textbf{0.82 $\pm$ 0.12} & \textbf{0.85 $\pm$ 0.08} & \textbf{0.76 $\pm$ 0.08} \\
\bottomrule
\end{tabular}

\vspace{4pt}

\textbf{Causality}\\
\begin{tabular}{lccccc}
\toprule
Method & Locations & Research Topic & Persons & Organizations & Organisms \\
\midrule
Input averaging & \textbf{0.55 $\pm$ 0.02} & 0.17 $\pm$ 0.04 & 0.19 $\pm$ 0.07 & 0.50 $\pm$ 0.09 & \textbf{0.15 $\pm$ 0.05} \\
SVM & 0.36 $\pm$ 0.03 & 0.13 $\pm$ 0.02 & 0.11 $\pm$ 0.03 & 0.46 $\pm$ 0.03 & 0.07 $\pm$ 0.01 \\
LHE & 0.53 $\pm$ 0.01 & \textbf{0.24 $\pm$ 0.07} & \textbf{0.28 $\pm$ 0.06} & \textbf{0.58 $\pm$ 0.03} & 0.12 $\pm$ 0.03 \\
\bottomrule
\end{tabular}
\caption{Accuracy and causality by domain for four LMs, evaluated using the prompt format shown in Figure~\ref{fig:prompt}. Scores are averaged across hierarchical levels. Standard deviations are computed across seeds after averaging each seed over hierarchical levels.}
\label{tab:acc_causality_all_models}

\end{table*}

\begin{table*}[t]
\centering
\small

\textbf{Llama 3.2 3B}\\
\textbf{Accuracy}\\
\begin{tabular}{lccccc}
\toprule
Method & Locations & Research Topic & Persons & Organizations & Organisms \\
\midrule
Input averaging & 0.44 $\pm$ 0.09 & 0.35 $\pm$ 0.16 & 0.62 $\pm$ 0.26 & 0.80 $\pm$ 0.18 & 0.47 $\pm$ 0.14 \\
SVM & 0.58 $\pm$ 0.08 & 0.35 $\pm$ 0.11 & 0.70 $\pm$ 0.18 & 0.91 $\pm$ 0.05 & 0.52 $\pm$ 0.08 \\
LHE & \textbf{0.78 $\pm$ 0.02} & \textbf{0.62 $\pm$ 0.05} & \textbf{0.99 $\pm$ 0.01} & \textbf{0.95 $\pm$ 0.03} & \textbf{0.81 $\pm$ 0.04} \\
\bottomrule
\end{tabular}

\vspace{4pt}

\textbf{Causality}\\
\begin{tabular}{lccccc}
\toprule
Method & Locations & Research Topic & Persons & Organizations & Organisms \\
\midrule
Input averaging & 0.74 $\pm$ 0.03 & 0.25 $\pm$ 0.05 & 0.52 $\pm$ 0.13 & 0.62 $\pm$ 0.10 & 0.47 $\pm$ 0.05 \\
SVM & 0.76 $\pm$ 0.03 & 0.28 $\pm$ 0.05 & 0.73 $\pm$ 0.03 & 0.74 $\pm$ 0.19 & 0.51 $\pm$ 0.07 \\
LHE & \textbf{0.83 $\pm$ 0.03} & \textbf{0.34 $\pm$ 0.06} & \textbf{0.82 $\pm$ 0.02} & \textbf{0.87 $\pm$ 0.02} & \textbf{0.72 $\pm$ 0.04} \\
\bottomrule
\end{tabular}

\vspace{10pt}

\textbf{Llama 3.1 8B}\\
\textbf{Accuracy}\\
\begin{tabular}{lccccc}
\toprule
Method & Locations & Research Topic & Persons & Organizations & Organisms \\
\midrule
Input averaging & 0.48 $\pm$ 0.09 & 0.39 $\pm$ 0.07 & 0.59 $\pm$ 0.15 & 0.64 $\pm$ 0.19 & 0.53 $\pm$ 0.08 \\
SVM & 0.61 $\pm$ 0.08 & 0.42 $\pm$ 0.07 & 0.65 $\pm$ 0.14 & 0.79 $\pm$ 0.05 & 0.57 $\pm$ 0.08 \\
LHE & \textbf{0.71 $\pm$ 0.02} & \textbf{0.59 $\pm$ 0.02} & \textbf{0.88 $\pm$ 0.03} & \textbf{0.98 $\pm$ 0.01} & \textbf{0.78 $\pm$ 0.03} \\
\bottomrule
\end{tabular}

\vspace{4pt}

\textbf{Causality}\\
\begin{tabular}{lccccc}
\toprule
Method & Locations & Research Topic & Persons & Organizations & Organisms \\
\midrule
Input averaging & 0.72 $\pm$ 0.01 & 0.24 $\pm$ 0.05 & 0.70 $\pm$ 0.10 & 0.54 $\pm$ 0.23 & 0.55 $\pm$ 0.09 \\
SVM & 0.73 $\pm$ 0.02 & 0.33 $\pm$ 0.09 & 0.82 $\pm$ 0.07 & 0.72 $\pm$ 0.08 & 0.56 $\pm$ 0.10 \\
LHE & \textbf{0.84 $\pm$ 0.02} & \textbf{0.57 $\pm$ 0.13} & \textbf{0.92 $\pm$ 0.04} & \textbf{0.87 $\pm$ 0.08} & \textbf{0.80 $\pm$ 0.05} \\
\bottomrule
\end{tabular}

\vspace{10pt}

\textbf{Qwen3 8B}\\
\textbf{Accuracy}\\
\begin{tabular}{lccccc}
\toprule
Method & Locations & Research Topic & Persons & Organizations & Organisms \\
\midrule
Input averaging & 0.44 $\pm$ 0.09 & 0.27 $\pm$ 0.05 & 0.59 $\pm$ 0.18 & 0.66 $\pm$ 0.26 & 0.57 $\pm$ 0.09 \\
SVM & 0.58 $\pm$ 0.06 & 0.32 $\pm$ 0.04 & 0.64 $\pm$ 0.16 & 0.87 $\pm$ 0.08 & 0.57 $\pm$ 0.08 \\
LHE & \textbf{0.78 $\pm$ 0.03} & \textbf{0.62 $\pm$ 0.09} & \textbf{0.95 $\pm$ 0.08} & \textbf{0.98 $\pm$ 0.02} & \textbf{0.79 $\pm$ 0.04} \\
\bottomrule
\end{tabular}

\vspace{4pt}

\textbf{Causality}\\
\begin{tabular}{lccccc}
\toprule
Method & Locations & Research Topic & Persons & Organizations & Organisms \\
\midrule
Input averaging & 0.78 $\pm$ 0.03 & 0.30 $\pm$ 0.09 & 0.67 $\pm$ 0.15 & 0.48 $\pm$ 0.06 & 0.46 $\pm$ 0.02 \\
SVM & 0.57 $\pm$ 0.06 & 0.13 $\pm$ 0.05 & 0.25 $\pm$ 0.02 & 0.53 $\pm$ 0.11 & 0.34 $\pm$ 0.04 \\
LHE & \textbf{0.80 $\pm$ 0.04} & \textbf{0.60 $\pm$ 0.03} & \textbf{0.77 $\pm$ 0.05} & \textbf{0.79 $\pm$ 0.06} & \textbf{0.63 $\pm$ 0.07} \\
\bottomrule
\end{tabular}

\vspace{10pt}

\textbf{Qwen3 14B}\\
\textbf{Accuracy}\\
\begin{tabular}{lccccc}
\toprule
Method & Locations & Research Topic & Persons & Organizations & Organisms \\
\midrule
Input averaging & 0.43 $\pm$ 0.08 & 0.34 $\pm$ 0.15 & 0.45 $\pm$ 0.10 & 0.65 $\pm$ 0.19 & 0.48 $\pm$ 0.18 \\
SVM & 0.63 $\pm$ 0.06 & 0.39 $\pm$ 0.11 & 0.61 $\pm$ 0.16 & 0.87 $\pm$ 0.02 & 0.52 $\pm$ 0.13 \\
LHE & \textbf{0.84 $\pm$ 0.02} & \textbf{0.61 $\pm$ 0.07} & \textbf{0.98 $\pm$ 0.02} & \textbf{0.98 $\pm$ 0.01} & \textbf{0.80 $\pm$ 0.03} \\
\bottomrule
\end{tabular}

\vspace{4pt}

\textbf{Causality}\\
\begin{tabular}{lccccc}
\toprule
Method & Locations & Research Topic & Persons & Organizations & Organisms \\
\midrule
Input averaging & \textbf{0.79 $\pm$ 0.01} & 0.28 $\pm$ 0.11 & 0.68 $\pm$ 0.06 & 0.55 $\pm$ 0.18 & \textbf{0.54 $\pm$ 0.09} \\
SVM & 0.45 $\pm$ 0.04 & 0.09 $\pm$ 0.03 & 0.25 $\pm$ 0.06 & 0.45 $\pm$ 0.12 & 0.32 $\pm$ 0.02 \\
LHE & 0.71 $\pm$ 0.03 & \textbf{0.38 $\pm$ 0.08} & \textbf{0.78 $\pm$ 0.12} & \textbf{0.75 $\pm$ 0.07} & 0.47 $\pm$ 0.07 \\
\bottomrule
\end{tabular}
\caption{Accuracy and causality by domain for four LMs, evaluated using the prompt format shown in Figure~\ref{fig:prompt_templates_fewshot}. Scores are averaged across hierarchical levels. Standard deviations are computed across seeds after averaging each seed over hierarchical levels.}
\label{tab:acc_causality_all_models_templates_fewshot}

\end{table*}

\subsection{Shuffled Test Parent Labels}
\label{sec:appendix_relation_swap_control}
Within each test relation, we randomly reassign parent labels so that every child receives a different parent from the same relation.
The reassignment preserves the parent-label counts whenever feasible in the input test set.
The shuffled evaluation reuses the fitted LHE maps and initializes parent concept vectors from correctly labeled pairs.
We average Accuracy over five seeds for each domain--depth setting and then macro-average over settings with results for both evaluations.

\begin{center}
\begin{minipage}{\columnwidth}
\captionsetup{type=table}
\centering
\small
\begin{tabular}{lrr}
\toprule
Model & Original & Shuffled \\
\midrule
Llama 3.2 3B & 0.84 & 0.018 \\
Llama 3.1 8B & 0.75 & 0.007 \\
Qwen3 8B     & 0.81 & 0.007 \\
Qwen3 14B    & 0.74 & 0.016 \\
\bottomrule
\end{tabular}
\caption{LHE decoding Accuracy with original and shuffled test parent labels.}
\label{tab:relation_swap_control}
\end{minipage}
\end{center}

\subsection{Target-Parent Rank}
\label{sec:appendix_target_parent_rank}
We use counterfactual examples in which the direct parent changes and all higher ancestors are shared by the original and counterfactual paths (App.~\ref{sec:appendix_counterfactual}).
For each query about the direct parent, we score the same answer choices with and without the LHE intervention.
The target parent's rank is one plus the number of choices with a strictly higher score.
Hits@$k$ is the fraction of queries in which the target has rank at most $k$.
We pool queries across the five domains separately for each model (Tbl.~\ref{tab:target_parent_rank}).
After editing, the target ranked first in 31.1--37.5\% of queries, depending on the model.

\begin{center}
\begin{minipage}{\columnwidth}
\captionsetup{type=table}
\centering
\small
\begin{tabular}{@{}lrrr@{}}
\toprule
Model & Mean rank & Hits@1 (\%) & Hits@3 (\%) \\
\midrule
Llama 3.2 3B & 4.60 $\to$ 2.77 & 1.6 $\to$ 34.5 & 40.0 $\to$ 71.9 \\
Llama 3.1 8B & 4.86 $\to$ 3.05 & 1.0 $\to$ 31.1 & 36.2 $\to$ 67.8 \\
Qwen3 8B     & 4.84 $\to$ 2.77 & 1.0 $\to$ 37.5 & 36.2 $\to$ 74.0 \\
Qwen3 14B    & 4.90 $\to$ 3.08 & 0.7 $\to$ 32.7 & 35.3 $\to$ 67.6 \\
\bottomrule
\end{tabular}
\caption{Target-parent rank among all answer choices before and after LHE editing.}
\label{tab:target_parent_rank}
\end{minipage}
\end{center}

\subsection{Interventions in Random Directions}
\label{sec:appendix_random_direction_control}
For each source--target concept pair, we sample an isotropic Gaussian vector and rescale it to match the norm of the LHE displacement $\mathbf{v}_{r,c'}-\mathbf{v}_{r,c}$.
We use the same random direction for that pair across edited layers and candidate-answer scoring.
We evaluate LHE and random-direction interventions on the same edits, using the same intervention coefficient and all-layer protocol (Sec.~\ref{sec:Experiments_setup}).
For each domain--depth setting, we average Causality over five seeds, then average across depths within each domain and equally across the five domains.
Tbl.~\ref{tab:random_direction_control} shows that random-direction Causality is substantially lower than LHE Causality for all four models.

\begin{table}[t]
\centering
\small
\begin{tabular}{lrr}
\toprule
Model & LHE & Random \\
\midrule
Llama 3.2 3B & 0.57 & 0.0034 \\
Llama 3.1 8B & 0.56 & 0.0014 \\
Qwen3 8B     & 0.47 & 0.0012 \\
Qwen3 14B    & 0.35 & 0.0136 \\
\bottomrule
\end{tabular}
\caption{Causality for LHE and norm-matched random-direction interventions across models. Scores are averaged over seeds, depths within each domain, and the five domains.}
\label{tab:random_direction_control}
\end{table}

\subsection{Intervention Locality}
\label{sec:appendix_intervention_locality}
To evaluate intervention locality, we pair each source-to-target edit with one held-out input belonging to a third concept within the same relation and hierarchy depth, excluding both the source and target concepts.
We retain only inputs answered correctly by greedy decoding without intervention.
We then apply each method's fixed source-to-target intervention direction to these inputs.
We define preservation as the fraction of inputs that remain correctly answered after intervention, and Side Effect as one minus this rate.
On Llama 3.1 8B, Causality/Side Effect is 0.56/0.21 for LHE, 0.35/0.19 for SVM, and 0.16/0.06 for Input Averaging.
Thus, compared with SVM, LHE gains 0.21 in Causality for a 0.02 increase in Side Effect.

\subsection{Comparison with a Nonlinear Decoder}
\label{sec:appendix_nonlinear_decoder}
As a supplementary comparison with a nonlinear decoder, we evaluated a two-layer MLP on Llama 3.1 8B.
The network consists of two linear layers ($4096 \to h \to 4096$, where $h$ is the hidden width), with GELU and dropout between them.
We trained a separate MLP for each domain--depth pair by minimizing mean one-minus-cosine similarity between predicted and target representations.
We used AdamW with a batch size of 64, weight decay of $10^{-4}$, and gradient clipping at a norm of 1.0.

We searched 18 configurations combining hidden widths $\{128, 256, 512\}$, learning rates $\{10^{-4}, 3\times10^{-4}, 10^{-3}\}$, and dropout rates $\{0.0, 0.1\}$.
Each candidate was trained for up to 150 epochs, with early stopping after 15 epochs without improvement in dev loss.
We evaluated each candidate on 12 domain--depth tasks with five seeds (42--46), averaged over seeds within each task, and then macro-averaged across tasks.
We selected one shared configuration by maximizing the harmonic mean of dev Accuracy and Causality.
The selected hidden width, learning rate, and dropout rate were 256, $10^{-3}$, and 0.1, respectively.
We then retrained each MLP on the full train+dev split for 66 epochs, determined by rounding the median best epoch across tasks and seeds for the selected configuration.

Using the same Accuracy evaluation as LHE, the MLP achieved 0.34, compared with 0.75 for LHE.
Thus, the MLP did not improve decoding accuracy in this comparison.

\section{Effect of Post-Training on Hierarchical Representations}
\label{sec:appendix_post_training}
To examine how post-training (e.g., SFT and RLHF) affects the representations of hierarchical relations, we compare matched base and Instruct models within the Llama 3 family.
Tbl.~\ref{tab:base_instruct} reports the results, averaged across all domains; the test set is kept identical across the four models.
We observe that Accuracy decreases only slightly in the Instruct models, by 0.04--0.05 points, whereas Causality decreases more substantially, by 0.07--0.10 points.
This suggests that hierarchical information remains linearly decodable to a similar extent after post-training, but its causal effect on next-token prediction becomes weaker.
In other words, post-training may change how the model uses the encoded hierarchical information.

\begin{table}[t]
\centering
\small
\begin{tabular}{lcc}
\toprule
Model & Accuracy & Causality \\
\midrule
Llama 3.2 3B          & 0.84 & 0.57 \\
Llama 3.2 3B Instruct & 0.80 & 0.47 \\
\midrule
Llama 3.1 8B          & 0.76 & 0.50 \\
Llama 3.1 8B Instruct & 0.71 & 0.43 \\
\bottomrule
\end{tabular}
\caption{Comparison of matched base and Instruct models within the Llama 3 family.
Scores are averaged across all domains, and the test set is kept identical across the four models.}
\label{tab:base_instruct}
\end{table}

\section{Test-Set Sensitivity to Rank and Layer Choices}
\label{sec:appendix_sweep}
\begin{figure*}[t]
\centering
\includegraphics[width=0.6\linewidth]{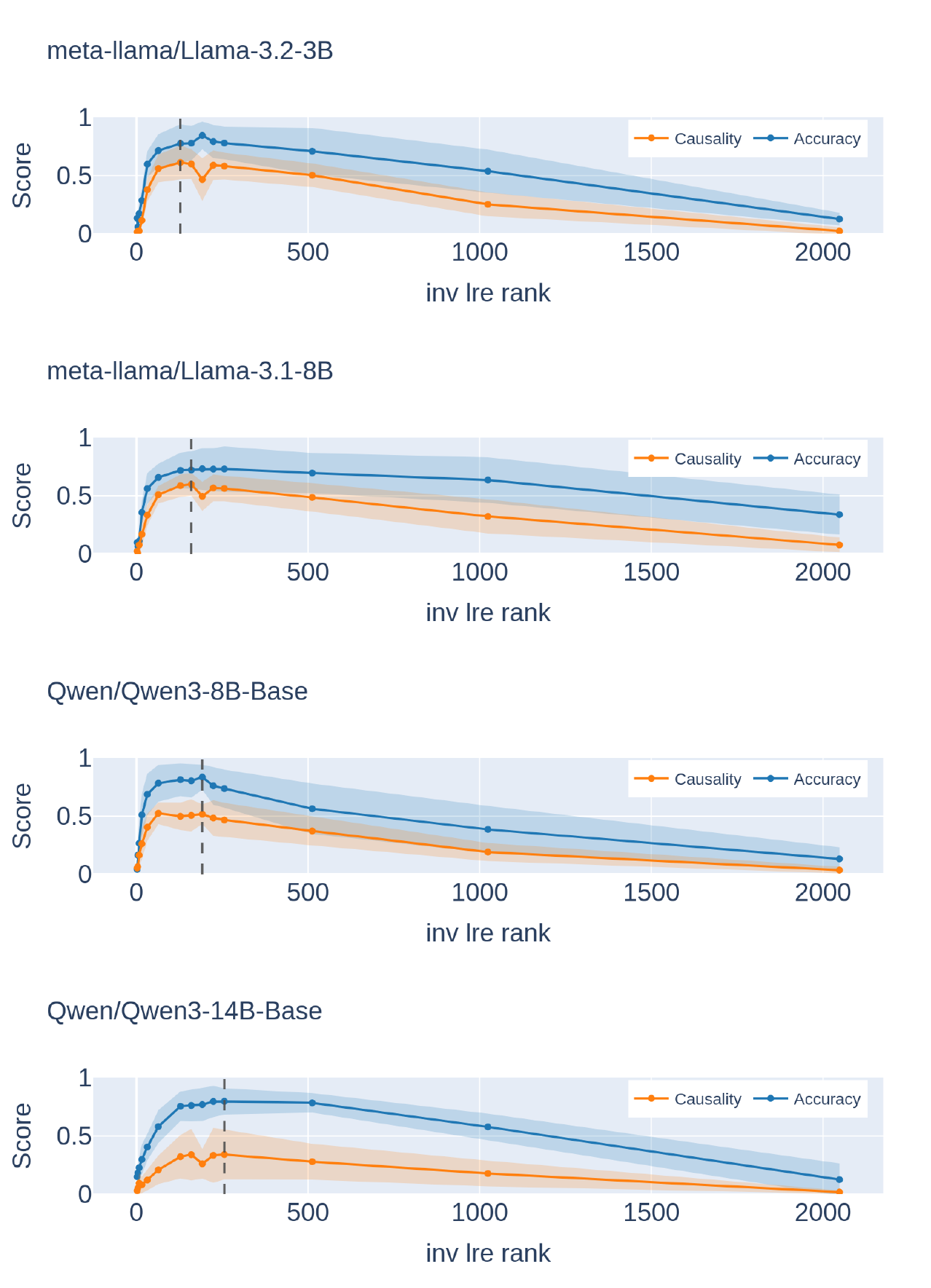}
\caption{
Post-hoc sweep of the truncation rank of $\mathbf{W}_r^{\dagger}$ for all models on the held-out test subtrees.
Scores are averaged across domains. The vertical dashed line in each panel marks the rank with the highest mean of test Accuracy and Causality.
}
\label{fig:sweep_rank_all_models}
\end{figure*}

\begin{figure*}[t]
\centering
\includegraphics[width=\linewidth]{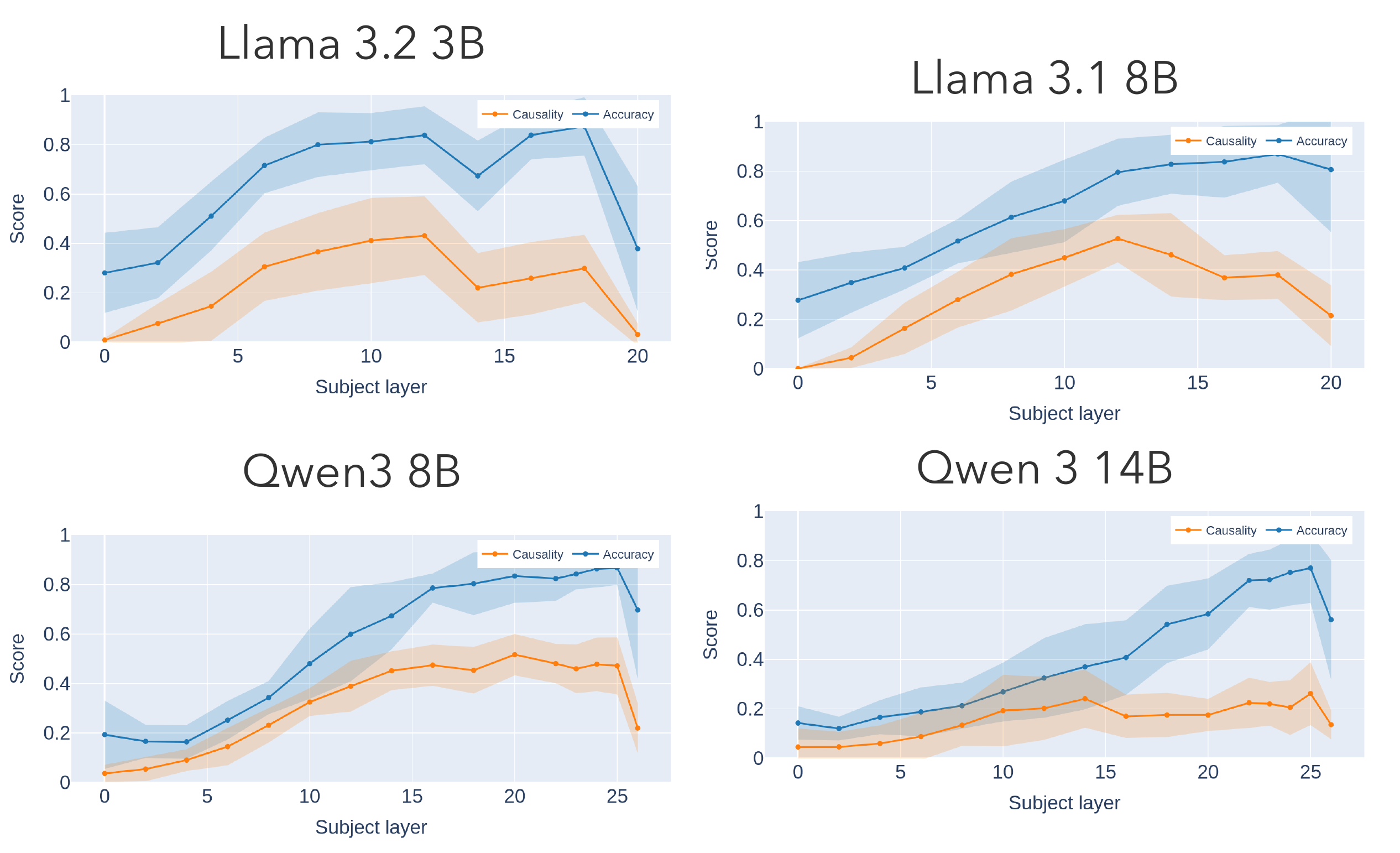}
\caption{
Post-hoc sensitivity of test Accuracy and Causality to the subject layer with the object layer fixed.
}
\label{fig:sweep_subj_all_models}
\end{figure*}

\begin{figure*}[t]
\centering
\includegraphics[width=\linewidth]{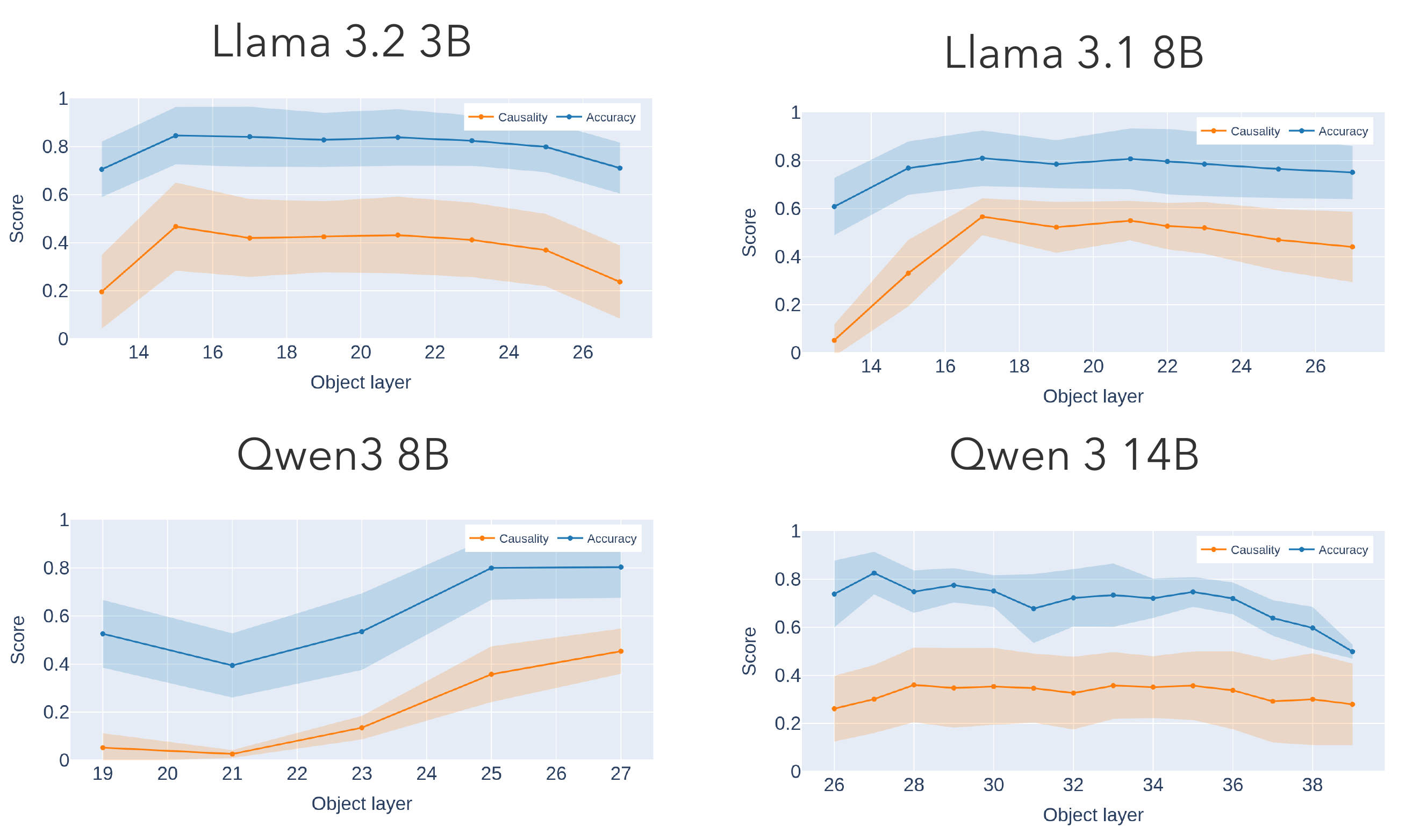}
\caption{
Post-hoc sensitivity of test Accuracy and Causality to the object layer with the subject layer fixed.
}
\label{fig:sweep_obj_all_models}
\end{figure*}

After selecting hyperparameters on dev (App.~\ref{sec:appendix_hyperparameter}), we conducted a post-hoc sensitivity analysis on the held-out test subtrees by varying the truncation rank of the pseudo-inverse (Fig.~\ref{fig:sweep_rank_all_models}), the subject layer (Fig.~\ref{fig:sweep_subj_all_models}), and the object layer (Fig.~\ref{fig:sweep_obj_all_models}).

\paragraph{Average Number of Children per Parent and Peak Rank.}
To examine the relationship between hierarchical branching and the rank at which performance peaks, we analyzed Llama 3.1 8B with subject layer 12 and object layer 21 across 12 relations from five domains.
For each relation, we compared the average number of children per parent with the peak rank, defined as the rank of the pseudo-inverse that maximizes each metric in the rank sweep, determined separately for Accuracy and Causality.
Spearman's rank correlation was $\rho=0.65$ for Accuracy and $\rho=0.12$ for Causality.
These results suggest that more children per parent may require a higher decoding rank.
Meanwhile, the association between the peak rank for Causality and the average number of children per parent was weak.

\subsection{Intervention Layer Ablation}
\label{sec:appendix_intervention_layers}
To examine how intervention depth affects Causality, we evaluate LHE on Llama 3.1 8B with the intervention defined in Sec.~\ref{sec:Experiments_setup} applied either to all layers or only to layers 1--10, 12--21, or 21--30.
These are the layers at which the concept vectors are injected, distinct from the subject and object layers used to estimate LHE.
Tbl.~\ref{tab:intervention_layers} reports the results.
Restricting the intervention to layers 1--10 yields Causality of 0.40, compared with 0.56 for all layers, whereas restricting it to layers 12--21 or 21--30 reduces Causality to 0.04 or 0.002, respectively.
Thus, the effectiveness of the intervention depends on where it is applied.
The drop for later layers is consistent with the possibility that relevant information has already propagated before the intervention~\citep{lrc}.

\begin{table}[t]
\centering
\small
\begin{tabular}{lr}
\toprule
Intervention layers & Causality \\
\midrule
All layers (default) & 0.56 \\
1--10 & 0.40 \\
12--21 & 0.04 \\
21--30 & 0.002 \\
\bottomrule
\end{tabular}
\caption{Effect of restricting the intervention to different layer ranges on Llama 3.1 8B.}
\label{tab:intervention_layers}
\end{table}

\section{PCA Visualizations}
\label{sec:appendix_pca}
\begin{figure*}[t]
\centering
\includegraphics[width=0.8\linewidth]{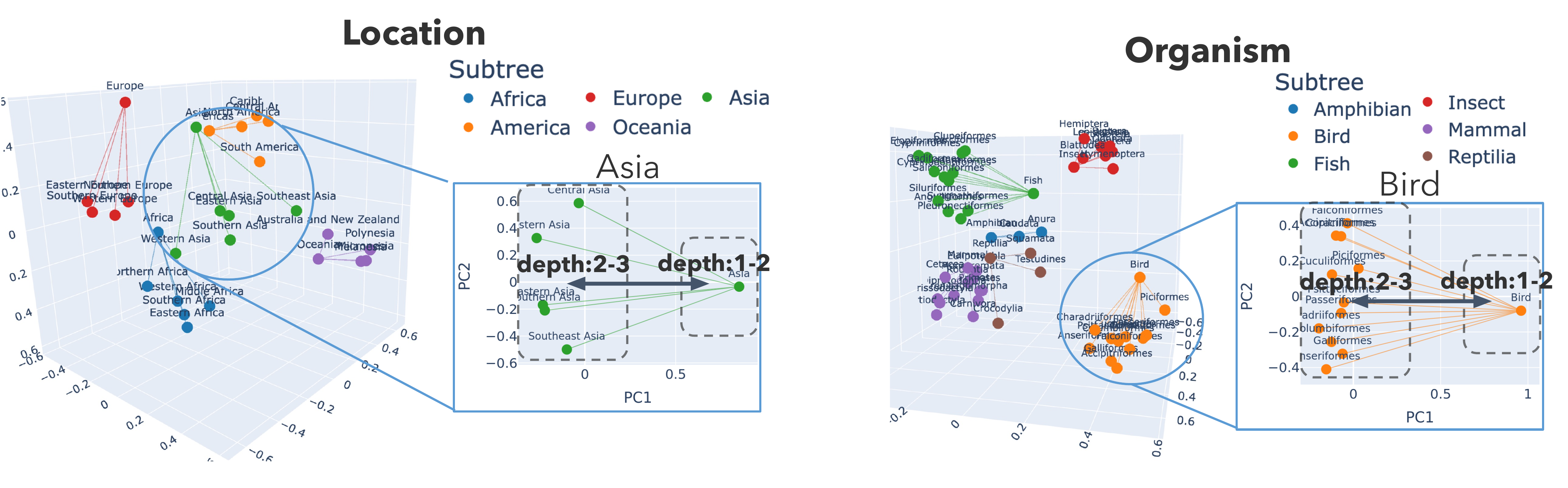}
\caption{PCA of concept vectors for Llama 3.1 8B in the Location and Organism domains. Concept vectors are obtained by applying a low-rank pseudo-inverse of the learned transformation to parent representations and averaging over examples grouped by parent concept.}
\label{fig:lrc_vec_pca}
\end{figure*}

\begin{figure*}[t]
\centering
\includegraphics[width=\linewidth]{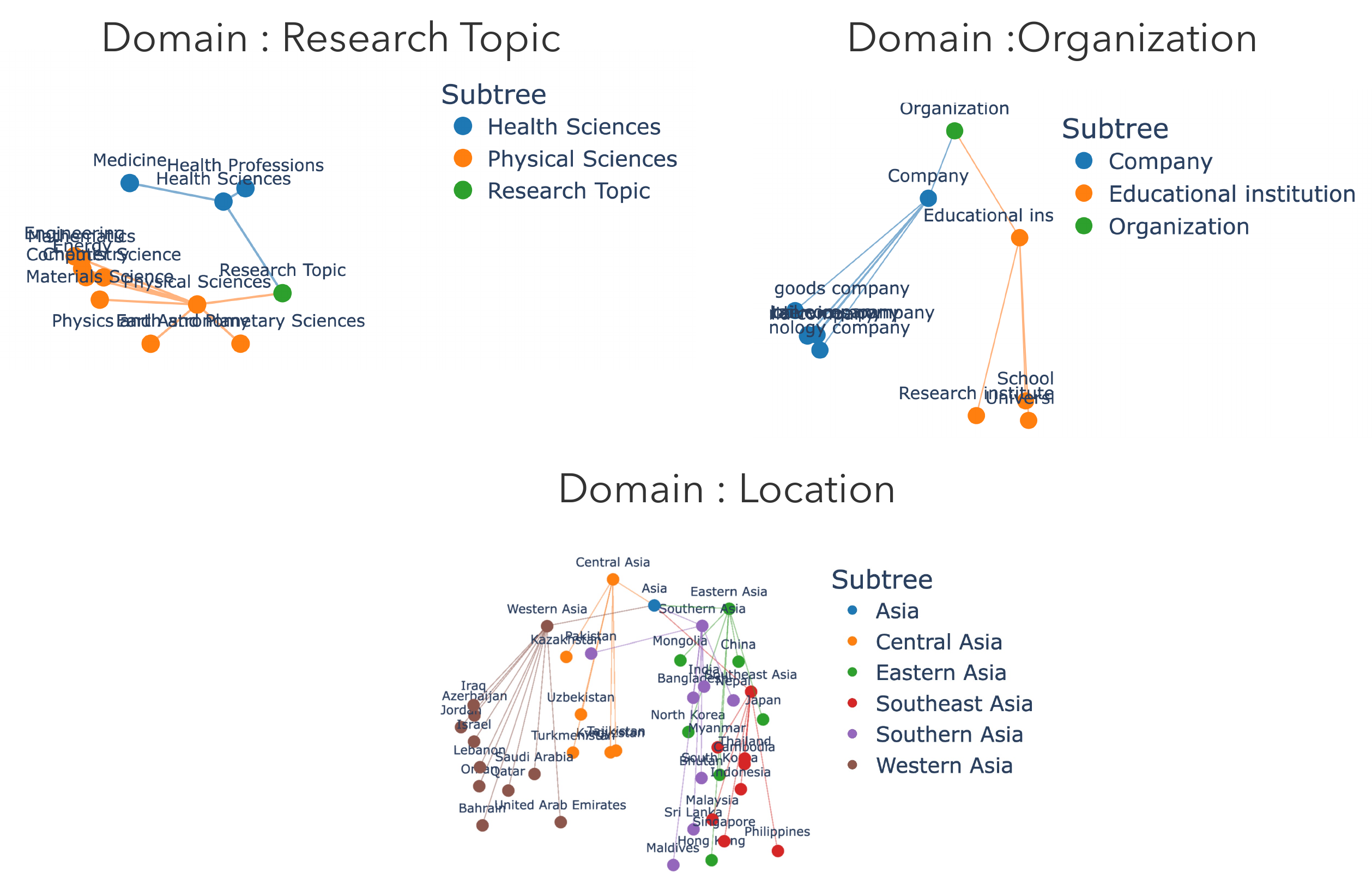}
\caption{PCA visualizations of the concept vectors for Llama 3.1 8B.}
\label{fig:pca_fig_appendix}
\end{figure*}

Figs.~\ref{fig:lrc_vec_pca} and~\ref{fig:pca_fig_appendix} show detailed PCA visualizations of the concept vectors obtained from Llama 3.1 8B.
Overall, the vectors form coherent clusters corresponding to subtrees.
For example, in the Location domain, vectors associated with concepts under Africa and Asia tend to group together (Fig.~\ref{fig:lrc_vec_pca}).
When we apply PCA within each subtree, the first two principal components often primarily reflect hierarchical depth: vectors at different depths are separated along these components.
This suggests a two-level organization in representation space: vectors first cluster by subtree (e.g., Asia), and then encode depth variation within the subtree.
We observe similar patterns in the other domains (Fig.~\ref{fig:pca_fig_appendix}), supporting the qualitative summary in Fig.~\ref{fig:pontie}.

\section{Experimental Details for the Topological Data Analysis}
\label{sec:appendix_tda}
\begin{figure*}[t]
\centering
\includegraphics[width=\linewidth]{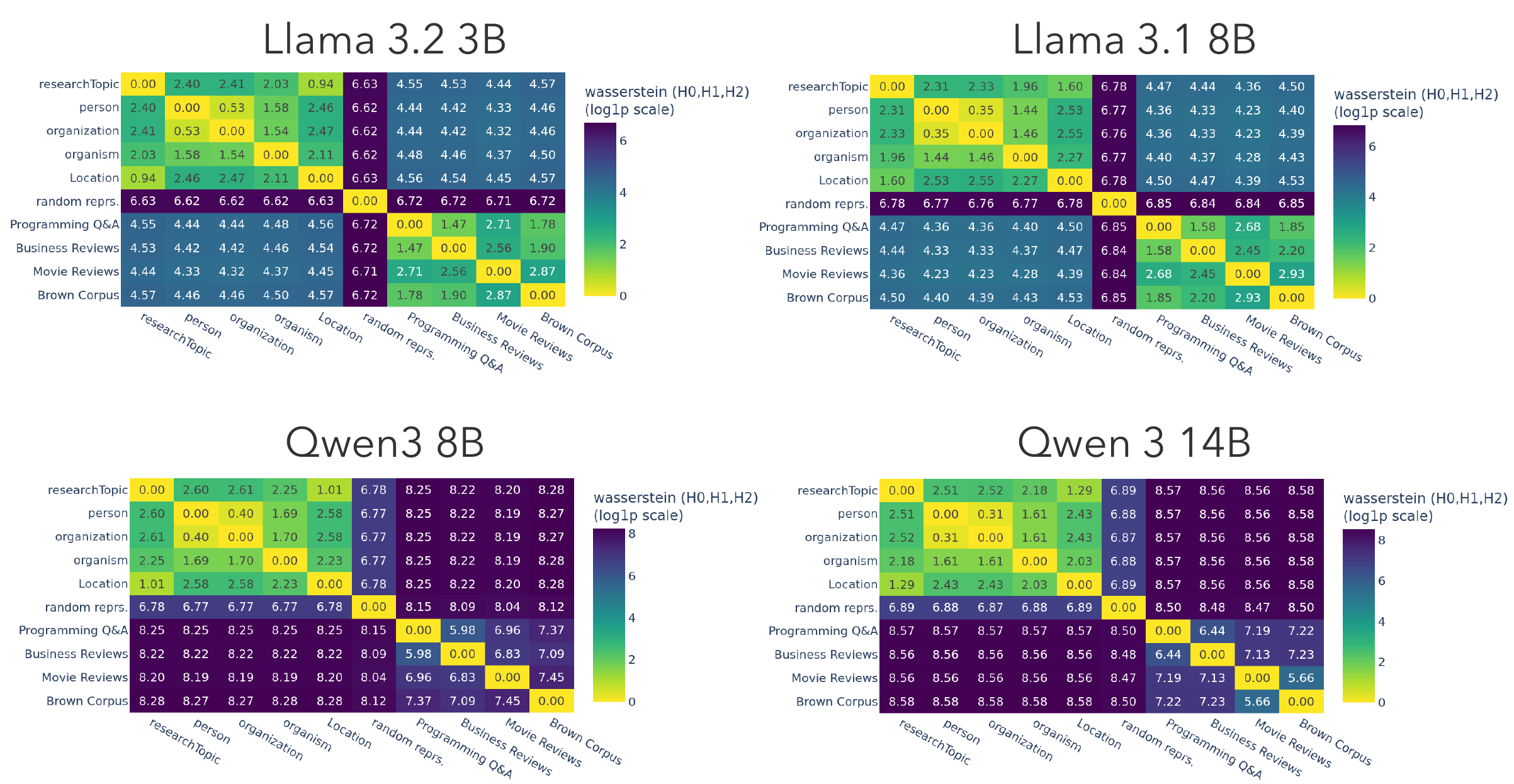}
\caption{Similarity matrices of LM representations for all models, based on persistent homology. Each entry shows the Wasserstein distance (computed on persistence diagrams derived from the concept vector point clouds; log1p scale) between a pair of domains/datasets. Smaller values indicate more similar representational structure.}
\label{fig:tda_four_model}
\end{figure*}

When two point clouds have aligned instances, correspondence-based measures such as Representational Similarity Analysis~\citep{RSA} or Centered Kernel Alignment~\citep{CKA} can compare them directly.
Entities do not align across our five domains, so we instead use persistent homology to obtain a correspondence-free description of each point cloud.
In our experiments, for each domain we treat the set of concept vectors as a point cloud, compute persistent homology under Euclidean distance, and obtain a persistence diagram.
We then quantify cross-domain similarity by computing the Wasserstein distance between persistence diagrams.
Intuitively, a persistence diagram summarizes how topological features of the point cloud (e.g., mergers of connected components, formation of loops) \emph{appear} (birth) and \emph{disappear} (death) as we gradually increase the radius $\epsilon$ of balls centered at each point; it represents these events as a multiset of birth--death pairs. Smaller distances between diagrams indicate more similar representational structure across domains.
As controls, we use random representations and four non-hierarchical corpora without hierarchy labels: Programming Q\&A~\citep{chandra_stackoverflow_dataset_2022}, Business Reviews~\citep{business_reviews}, Movie Reviews~\citep{movie_review}, and the Brown Corpus~\citep{brown_corpus}.
We randomly sample nouns from each corpus, extract their representations directly from the subject layer, and apply the same procedure.
To reduce sample-size bias, each baseline uses the median sample size across the hierarchical domains.
The results for all models obtained in these experiments are shown in Fig.~\ref{fig:tda_four_model}.

\section{Details of the Counterfactual Editing Experiment}
\label{sec:appendix_counterfactual}

\paragraph{Counterfactual data.}
For each model, we start from its LM-filtered hierarchy (App.~\ref{sec:appendix_data}).
For each leaf item, we replace the direct parent with a counterfactual parent from the same domain hierarchy and construct the counterfactual path upward from that parent; the Up-to-$k$hop condition selects the counterfactual parent such that the counterfactual path diverges from the original path for exactly $k$ hops (Fig.~\ref{fig:counterfactual_conditions}).
Persons, Organizations, and Organisms are evaluated up to 2 hops, and Locations and Research Topic up to 3 hops.
To balance the evaluation set, we sample at most five items per source concept and exclude items that overlap with the data used to estimate the intervention directions.
Tbl.~\ref{tab:counterfactual_data_num} reports the number of evaluated leaf items; within a model and domain, all conditions use the same items.
Because the LM-filtered hierarchies differ across models, cross-model differences are not computed on identical samples.

\begin{figure*}[t]
\centering
\includegraphics[width=\linewidth]{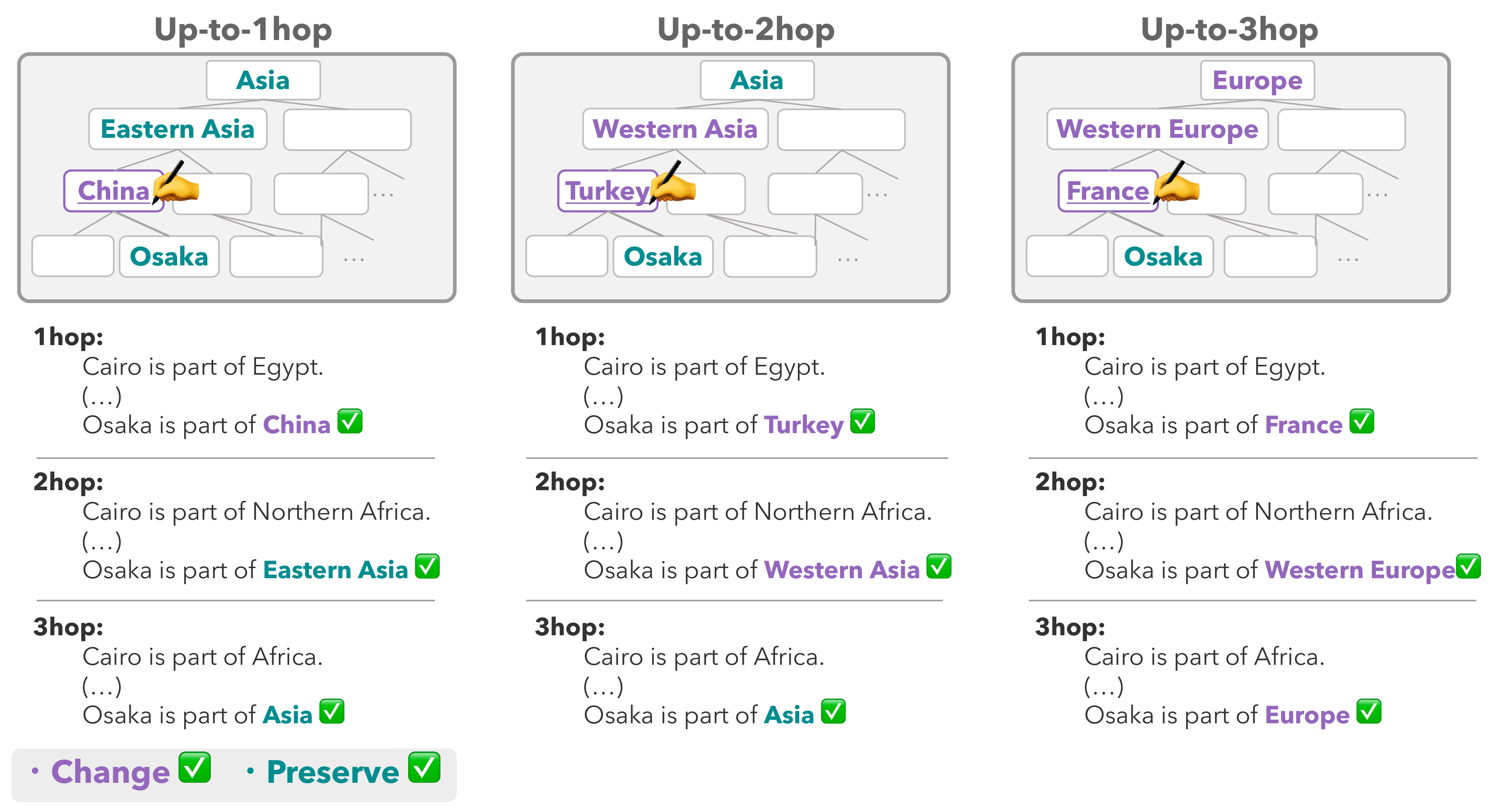}
\caption{Concrete illustration of the counterfactual conditions and hop-specific prompting. Each column shows an Up-to-$k$hop condition created by editing only the direct parent. For each distance $h$, the few-shot demonstrations use $h$-hop ancestors as labels, thereby specifying which ancestor the unchanged query template should predict. Purple and teal denote Change and Preserve, respectively.}
\label{fig:counterfactual_prompt_example}
\end{figure*}

\paragraph{Evaluation protocol.}
Prompts follow the multiple-choice QA format of Fig.~\ref{fig:prompt} with eight few-shot examples.
To ask for the ancestor $h$ hops from the leaf, we construct a separate few-shot prompt for each $h$: its demonstrations map leaf items to their $h$-hop ancestors, while the query uses the same relation template (Fig.~\ref{fig:counterfactual_prompt_example}).
This prompt-level specification of the requested ancestor distance is distinct from the Up-to-$k$hop condition, which is determined by the counterfactual parent and path.

For each ancestor distance, we query the model separately and score every candidate label $y$ by the minimum token probability $s(y\mid x)=\min_{t}p(y_t\mid x,y_{<t})$, as in the Causality evaluation.
At an ancestor distance requiring Change, with original label $y_{\mathrm{orig}}$ and counterfactual label $y_{\mathrm{cf}}$, the edit is successful if
\[
s(y_{\mathrm{cf}}\mid x)>s(y_{\mathrm{orig}}\mid x).
\]
At an ancestor distance requiring Preserve, with gold label $y_{\mathrm{gold}}$ and candidate set $\mathcal{C}$, the edit is successful if
\[
s(y_{\mathrm{gold}}\mid x)>\max_{y\in\mathcal{C},\,y\neq y_{\mathrm{gold}}}s(y\mid x).
\]
We also compute all metrics for a no-edit baseline that evaluates the model without any intervention.
Without editing, Change rates are near zero in most cells (at most 0.23 in any cell), while Preserve rates range from 0.59 to 1.00, so post-edit Change success is attributable to the intervention.

\paragraph{Intervention details.}
The intervention procedure is the same as in the Causality evaluation (Sec.~\ref{sec:Experiments_setup}): we edit the final-token representation of the leaf at every layer by subtracting the source concept vector and adding the target concept vector, scaled by $\beta$ times the hidden-state norm.
The subject layer, object layer, rank, and $\beta$ are identical to the main experiments (Tbl.~\ref{tab:lrc_hyperparameter}).
LHE directions are obtained from the map of the deepest depth transition in each domain (depth $3\to2$ for Persons, Organisms, and Organizations; depth $4\to3$ for Locations and Research Topic), using eight training samples per parent concept.
The SVM baseline learns direction vectors with a linear SVM on the same training data, and the Input Averaging baseline uses the mean input representation associated with each parent as its direction.
Tbl.~\ref{tab:counterfactual_hopwise_all} reports the LHE results for all models, domains, conditions, and ancestor distances.
The corresponding Input Averaging and SVM results are reported in Tables~\ref{tab:cf-khop-input-avg} and~\ref{tab:cf-khop-svm}, respectively.
Tbl.~\ref{tab:counterfactual_method} summarizes these results using an unweighted macro-average over the available cells across domains, conditions, and ancestor distances, so that each cell contributes equally regardless of its number of examples.

\paragraph{Successful examples.}
Tbl.~\ref{tab:counterfactual_success_examples} presents one successful LHE edit for each condition for Llama 3.1 8B on Locations.
In each case, only the direct-parent relation is edited, while the outcomes are evaluated at every ancestor distance.

\begin{table*}[t]
\centering
\begin{tabularx}{\textwidth}{lXX}
\toprule
Condition & Original path & Counterfactual path and hop outcome \\
\midrule
Up-to-1hop &
Bukit Panjang $\rightarrow$ Singapore\newline
$\rightarrow$ Southeast Asia $\rightarrow$ Asia &
Bukit Panjang $\rightarrow$ Philippines (C, $\checkmark$)\newline
$\rightarrow$ Southeast Asia (P, $\checkmark$) $\rightarrow$ Asia (P, $\checkmark$) \\
\midrule
Up-to-2hop &
Salman Pak $\rightarrow$ Iraq\newline
$\rightarrow$ Western Asia $\rightarrow$ Asia &
Salman Pak $\rightarrow$ Singapore (C, $\checkmark$)\newline
$\rightarrow$ Southeast Asia (C, $\checkmark$) $\rightarrow$ Asia (P, $\checkmark$) \\
\midrule
Up-to-3hop &
Mampong $\rightarrow$ Ghana\newline
$\rightarrow$ Western Africa $\rightarrow$ Africa &
Mampong $\rightarrow$ Philippines (C, $\checkmark$)\newline
$\rightarrow$ Southeast Asia (C, $\checkmark$) $\rightarrow$ Asia (C, $\checkmark$) \\
\bottomrule
\end{tabularx}
\caption{Successful LHE counterfactual examples for Llama 3.1 8B on Locations. C and P denote Change and Preserve, respectively; a check mark denotes success at that ancestor distance.}
\label{tab:counterfactual_success_examples}
\end{table*}

\paragraph{Failure cases.}
For Llama 3.1 8B with LHE across the five domains, 1,400 of 4,035 cases requiring Change succeed (0.347).
Among the 2,635 failed Change cases, the top-ranked candidate is identical before and after the intervention in 2,272 cases (86.2\%), and the original label remains top-ranked after the intervention in 2,193 cases (83.2\%).
The dominant failure mode is therefore not moving to a wrong label but not moving at all: for example, an intervention intended to change \emph{Tokyo} $\to$ \emph{Japan} into \emph{Tokyo} $\to$ \emph{France} leaves \emph{Japan} as the top prediction.
This analysis has so far been conducted only for Llama 3.1 8B, and we do not claim that the same proportions hold for the other models.

\paragraph{Relation to entity popularity.}
We test whether intervention success is related to entity popularity, using aggregated 2023 Wikipedia page views only as a proxy for pretraining frequency~\citep{mallen-etal-2023-trust}.
Across all four LMs, Change success correlates positively with the combined log-scaled popularity of the child and counterfactual parent: Pearson $r=0.216$ for Llama 3.1 8B, $0.197$ for Llama 3.2 3B, $0.193$ for Qwen3 8B, and $0.125$ for Qwen3 14B (all $p<.001$).
These correlations are weak and observational.
They are nevertheless consistent with the frequency-related pattern reported by \citet{merullo2025linear}, who find that subject--object co-occurrences in the pretraining corpus predict the quality of linear relation representations.

\section{Use of AI Assistants}
We used OpenAI Codex to assist with improving the clarity, organization, and wording of the manuscript. We also used GPT-5 to generate candidate entities for the Person and Organization datasets. As described in Appendix~\ref{sec:appendix_data}, we retained only candidates corresponding to real-world entities in Wikidata. All AI-assisted content was reviewed by the authors, who take full responsibility for the manuscript, data construction procedure, experimental results, claims, and citations. Codex was not used to generate experimental results or make research decisions.

\begin{table}[t]
\centering
\small
\setlength{\tabcolsep}{4pt}
\begin{tabular}{lrrrrr}
\toprule
Model & Loc. & Topic & Pers. & Org. & Orgm. \\
\midrule
Llama 3.2 3B & 293 & 119 & 25 & 30 & 46 \\
Llama 3.1 8B & 361 & 250 & 28 & 40 & 55 \\
Qwen3 8B & 354 & 221 & 30 & 35 & 51 \\
Qwen3 14B & 377 & 255 & 30 & 35 & 50 \\
\bottomrule
\end{tabular}
\caption{Number of evaluated leaf items per model and domain in the counterfactual editing experiment. Within a model and domain, every condition uses the same items.}
\label{tab:counterfactual_data_num}
\end{table}

\begin{table}[t]
\centering
\small
\setlength{\tabcolsep}{4pt}
\begin{tabular}{lcccccc}
\toprule
& \multicolumn{2}{c}{Input avg.} & \multicolumn{2}{c}{SVM} & \multicolumn{2}{c}{LHE} \\
\cmidrule(lr){2-3} \cmidrule(lr){4-5} \cmidrule(lr){6-7}
Model & Chg. & Prs. & Chg. & Prs. & Chg. & Prs. \\
\midrule
Llama 3.2 3B & 0.24 & \textbf{0.71} & 0.30 & 0.57 & \textbf{0.45} & 0.66 \\
Llama 3.1 8B & 0.22 & \textbf{0.87} & 0.28 & 0.83 & \textbf{0.43} & 0.86 \\
Qwen3 8B & 0.20 & \textbf{0.90} & 0.21 & 0.82 & \textbf{0.41} & 0.87 \\
Qwen3 14B & 0.26 & \textbf{0.88} & 0.16 & 0.76 & \textbf{0.30} & 0.78 \\
\bottomrule
\end{tabular}
\caption{Post-intervention success rates for cases requiring Change (Chg.) and Preserve (Prs.) for each intervention method. Scores average all available cells across domains, conditions, and ancestor distances with equal weight (21 Change and 9 Preserve cells per model). Bold marks the best method per model and metric.}
\label{tab:counterfactual_method}
\end{table}

\begin{table*}[t]
\centering
\small
\begin{tabular}{lllccc}
\toprule
Model & Domain & Condition & 1-hop & 2-hop & 3-hop \\
\midrule
\multirow{12}{*}{Llama 3.2 3B} & \multirow{3}{*}{Locations} & Up-to-1hop & 0.40 (C) & 0.75 (P) & 0.65 (P) \\
 &  & Up-to-2hop & 0.20 (C) & 0.52 (C) & 0.59 (P) \\
 &  & Up-to-3hop & 0.21 (C) & 0.54 (C) & 0.63 (C) \\
\cmidrule(lr){2-6}
 & \multirow{3}{*}{Research Topic} & Up-to-1hop & 0.15 (C) & 0.76 (P) & 0.61 (P) \\
 &  & Up-to-2hop & 0.00 (C) & 0.13 (C) & 0.61 (P) \\
 &  & Up-to-3hop & 0.00 (C) & 0.46 (C) & 0.43 (C) \\
\cmidrule(lr){2-6}
 & \multirow{2}{*}{Persons} & Up-to-1hop & 0.80 (C) & 0.44 (P) & -- \\
 &  & Up-to-2hop & 0.12 (C) & 0.88 (C) & -- \\
\cmidrule(lr){2-6}
 & \multirow{2}{*}{Organizations} & Up-to-1hop & 0.70 (C) & 0.90 (P) & -- \\
 &  & Up-to-2hop & 0.07 (C) & 0.67 (C) & -- \\
\cmidrule(lr){2-6}
 & \multirow{2}{*}{Organisms} & Up-to-1hop & 0.98 (C) & 0.65 (P) & -- \\
 &  & Up-to-2hop & 0.61 (C) & 0.89 (C) & -- \\
\midrule
\multirow{12}{*}{Llama 3.1 8B} & \multirow{3}{*}{Locations} & Up-to-1hop & 0.42 (C) & 0.85 (P) & 0.85 (P) \\
 &  & Up-to-2hop & 0.32 (C) & 0.55 (C) & 0.85 (P) \\
 &  & Up-to-3hop & 0.30 (C) & 0.56 (C) & 0.63 (C) \\
\cmidrule(lr){2-6}
 & \multirow{3}{*}{Research Topic} & Up-to-1hop & 0.08 (C) & 0.80 (P) & 0.74 (P) \\
 &  & Up-to-2hop & 0.00 (C) & 0.20 (C) & 0.76 (P) \\
 &  & Up-to-3hop & 0.00 (C) & 0.13 (C) & 0.17 (C) \\
\cmidrule(lr){2-6}
 & \multirow{2}{*}{Persons} & Up-to-1hop & 0.64 (C) & 0.96 (P) & -- \\
 &  & Up-to-2hop & 0.04 (C) & 0.71 (C) & -- \\
\cmidrule(lr){2-6}
 & \multirow{2}{*}{Organizations} & Up-to-1hop & 0.65 (C) & 1.00 (P) & -- \\
 &  & Up-to-2hop & 0.17 (C) & 0.78 (C) & -- \\
\cmidrule(lr){2-6}
 & \multirow{2}{*}{Organisms} & Up-to-1hop & 0.85 (C) & 0.96 (P) & -- \\
 &  & Up-to-2hop & 0.84 (C) & 0.98 (C) & -- \\
\midrule
\multirow{12}{*}{Qwen3 8B} & \multirow{3}{*}{Locations} & Up-to-1hop & 0.50 (C) & 0.75 (P) & 0.82 (P) \\
 &  & Up-to-2hop & 0.16 (C) & 0.59 (C) & 0.79 (P) \\
 &  & Up-to-3hop & 0.11 (C) & 0.60 (C) & 0.62 (C) \\
\cmidrule(lr){2-6}
 & \multirow{3}{*}{Research Topic} & Up-to-1hop & 0.14 (C) & 0.87 (P) & 0.91 (P) \\
 &  & Up-to-2hop & 0.00 (C) & 0.23 (C) & 0.90 (P) \\
 &  & Up-to-3hop & 0.00 (C) & 0.11 (C) & 0.12 (C) \\
\cmidrule(lr){2-6}
 & \multirow{2}{*}{Persons} & Up-to-1hop & 0.83 (C) & 0.90 (P) & -- \\
 &  & Up-to-2hop & 0.10 (C) & 0.80 (C) & -- \\
\cmidrule(lr){2-6}
 & \multirow{2}{*}{Organizations} & Up-to-1hop & 0.77 (C) & 0.91 (P) & -- \\
 &  & Up-to-2hop & 0.11 (C) & 0.66 (C) & -- \\
\cmidrule(lr){2-6}
 & \multirow{2}{*}{Organisms} & Up-to-1hop & 0.84 (C) & 0.96 (P) & -- \\
 &  & Up-to-2hop & 0.47 (C) & 0.92 (C) & -- \\
\midrule
\multirow{12}{*}{Qwen3 14B} & \multirow{3}{*}{Locations} & Up-to-1hop & 0.50 (C) & 0.70 (P) & 0.66 (P) \\
 &  & Up-to-2hop & 0.06 (C) & 0.46 (C) & 0.60 (P) \\
 &  & Up-to-3hop & 0.02 (C) & 0.45 (C) & 0.51 (C) \\
\cmidrule(lr){2-6}
 & \multirow{3}{*}{Research Topic} & Up-to-1hop & 0.11 (C) & 0.85 (P) & 0.87 (P) \\
 &  & Up-to-2hop & 0.00 (C) & 0.15 (C) & 0.88 (P) \\
 &  & Up-to-3hop & 0.00 (C) & 0.12 (C) & 0.16 (C) \\
\cmidrule(lr){2-6}
 & \multirow{2}{*}{Persons} & Up-to-1hop & 0.73 (C) & 0.70 (P) & -- \\
 &  & Up-to-2hop & 0.00 (C) & 0.57 (C) & -- \\
\cmidrule(lr){2-6}
 & \multirow{2}{*}{Organizations} & Up-to-1hop & 0.71 (C) & 1.00 (P) & -- \\
 &  & Up-to-2hop & 0.00 (C) & 0.11 (C) & -- \\
\cmidrule(lr){2-6}
 & \multirow{2}{*}{Organisms} & Up-to-1hop & 0.82 (C) & 0.76 (P) & -- \\
 &  & Up-to-2hop & 0.02 (C) & 0.80 (C) & -- \\
\bottomrule
\end{tabular}
\caption{Post-intervention success rates of LHE by condition and ancestor distance for all models and domains. Each score is followed by its expected behavior (C: Change; P: Preserve). Persons, Organizations, and Organisms are evaluated up to 2 hops; Locations and Research Topic up to 3 hops.}
\label{tab:counterfactual_hopwise_all}
\end{table*}

\newcommand{\cfBaselineTable}[3]{%
\begin{table*}[t]
\centering
\small
\begin{tabular}{lllccc}
\toprule
Model & Domain & Condition & 1-hop & 2-hop & 3-hop \\
\midrule
#3
\bottomrule
\end{tabular}
\caption{Post-intervention success rates of #1 by condition and ancestor distance for all models and domains. Each score is followed by its expected behavior (C: Change; P: Preserve). Persons, Organizations, and Organisms are evaluated up to 2 hops; Locations and Research Topic up to 3 hops.}
\label{tab:cf-khop-#2}
\end{table*}%
}

\cfBaselineTable{Input Averaging}{input-avg}{%
\multirow{12}{*}{Llama 3.2 3B} & \multirow{3}{*}{Locations} & Up-to-1hop & 0.22 (C) & 0.81 (P) & 0.66 (P) \\
 &  & Up-to-2hop & 0.06 (C) & 0.30 (C) & 0.67 (P) \\
 &  & Up-to-3hop & 0.04 (C) & 0.31 (C) & 0.38 (C) \\
\cmidrule(lr){2-6}
 & \multirow{3}{*}{Research Topic} & Up-to-1hop & 0.06 (C) & 0.75 (P) & 0.62 (P) \\
 &  & Up-to-2hop & 0.00 (C) & 0.08 (C) & 0.61 (P) \\
 &  & Up-to-3hop & 0.00 (C) & 0.33 (C) & 0.33 (C) \\
\cmidrule(lr){2-6}
 & \multirow{2}{*}{Persons} & Up-to-1hop & 0.40 (C) & 0.56 (P) & -- \\
 &  & Up-to-2hop & 0.00 (C) & 0.52 (C) & -- \\
\cmidrule(lr){2-6}
 & \multirow{2}{*}{Organizations} & Up-to-1hop & 0.17 (C) & 0.87 (P) & -- \\
 &  & Up-to-2hop & 0.00 (C) & 0.30 (C) & -- \\
\cmidrule(lr){2-6}
 & \multirow{2}{*}{Organisms} & Up-to-1hop & 0.57 (C) & 0.85 (P) & -- \\
 &  & Up-to-2hop & 0.33 (C) & 0.76 (C) & -- \\
\midrule
\multirow{12}{*}{Llama 3.1 8B} & \multirow{3}{*}{Locations} & Up-to-1hop & 0.29 (C) & 0.87 (P) & 0.87 (P) \\
 &  & Up-to-2hop & 0.12 (C) & 0.31 (C) & 0.86 (P) \\
 &  & Up-to-3hop & 0.12 (C) & 0.28 (C) & 0.35 (C) \\
\cmidrule(lr){2-6}
 & \multirow{3}{*}{Research Topic} & Up-to-1hop & 0.04 (C) & 0.79 (P) & 0.72 (P) \\
 &  & Up-to-2hop & 0.00 (C) & 0.10 (C) & 0.71 (P) \\
 &  & Up-to-3hop & 0.00 (C) & 0.06 (C) & 0.08 (C) \\
\cmidrule(lr){2-6}
 & \multirow{2}{*}{Persons} & Up-to-1hop & 0.39 (C) & 1.00 (P) & -- \\
 &  & Up-to-2hop & 0.00 (C) & 0.36 (C) & -- \\
\cmidrule(lr){2-6}
 & \multirow{2}{*}{Organizations} & Up-to-1hop & 0.12 (C) & 1.00 (P) & -- \\
 &  & Up-to-2hop & 0.00 (C) & 0.17 (C) & -- \\
\cmidrule(lr){2-6}
 & \multirow{2}{*}{Organisms} & Up-to-1hop & 0.60 (C) & 0.98 (P) & -- \\
 &  & Up-to-2hop & 0.47 (C) & 0.75 (C) & -- \\
\midrule
\multirow{12}{*}{Qwen3 8B} & \multirow{3}{*}{Locations} & Up-to-1hop & 0.30 (C) & 0.83 (P) & 0.89 (P) \\
 &  & Up-to-2hop & 0.08 (C) & 0.26 (C) & 0.87 (P) \\
 &  & Up-to-3hop & 0.06 (C) & 0.22 (C) & 0.36 (C) \\
\cmidrule(lr){2-6}
 & \multirow{3}{*}{Research Topic} & Up-to-1hop & 0.03 (C) & 0.82 (P) & 0.91 (P) \\
 &  & Up-to-2hop & 0.00 (C) & 0.09 (C) & 0.90 (P) \\
 &  & Up-to-3hop & 0.00 (C) & 0.01 (C) & 0.03 (C) \\
\cmidrule(lr){2-6}
 & \multirow{2}{*}{Persons} & Up-to-1hop & 0.47 (C) & 0.93 (P) & -- \\
 &  & Up-to-2hop & 0.00 (C) & 0.30 (C) & -- \\
\cmidrule(lr){2-6}
 & \multirow{2}{*}{Organizations} & Up-to-1hop & 0.14 (C) & 1.00 (P) & -- \\
 &  & Up-to-2hop & 0.03 (C) & 0.14 (C) & -- \\
\cmidrule(lr){2-6}
 & \multirow{2}{*}{Organisms} & Up-to-1hop & 0.71 (C) & 0.96 (P) & -- \\
 &  & Up-to-2hop & 0.27 (C) & 0.67 (C) & -- \\
\midrule
\multirow{12}{*}{Qwen3 14B} & \multirow{3}{*}{Locations} & Up-to-1hop & 0.40 (C) & 0.81 (P) & 0.83 (P) \\
 &  & Up-to-2hop & 0.11 (C) & 0.37 (C) & 0.81 (P) \\
 &  & Up-to-3hop & 0.06 (C) & 0.39 (C) & 0.43 (C) \\
\cmidrule(lr){2-6}
 & \multirow{3}{*}{Research Topic} & Up-to-1hop & 0.02 (C) & 0.84 (P) & 0.87 (P) \\
 &  & Up-to-2hop & 0.00 (C) & 0.05 (C) & 0.87 (P) \\
 &  & Up-to-3hop & 0.00 (C) & 0.03 (C) & 0.09 (C) \\
\cmidrule(lr){2-6}
 & \multirow{2}{*}{Persons} & Up-to-1hop & 0.47 (C) & 0.97 (P) & -- \\
 &  & Up-to-2hop & 0.03 (C) & 0.43 (C) & -- \\
\cmidrule(lr){2-6}
 & \multirow{2}{*}{Organizations} & Up-to-1hop & 0.34 (C) & 1.00 (P) & -- \\
 &  & Up-to-2hop & 0.00 (C) & 0.37 (C) & -- \\
\cmidrule(lr){2-6}
 & \multirow{2}{*}{Organisms} & Up-to-1hop & 0.72 (C) & 0.96 (P) & -- \\
 &  & Up-to-2hop & 0.28 (C) & 0.82 (C) & -- \\
}

\cfBaselineTable{SVM}{svm}{%
\multirow{12}{*}{Llama 3.2 3B} & \multirow{3}{*}{Locations} & Up-to-1hop & 0.33 (C) & 0.72 (P) & 0.55 (P) \\
 &  & Up-to-2hop & 0.07 (C) & 0.35 (C) & 0.52 (P) \\
 &  & Up-to-3hop & 0.03 (C) & 0.37 (C) & 0.41 (C) \\
\cmidrule(lr){2-6}
 & \multirow{3}{*}{Research Topic} & Up-to-1hop & 0.07 (C) & 0.76 (P) & 0.61 (P) \\
 &  & Up-to-2hop & 0.00 (C) & 0.09 (C) & 0.59 (P) \\
 &  & Up-to-3hop & 0.00 (C) & 0.33 (C) & 0.30 (C) \\
\cmidrule(lr){2-6}
 & \multirow{2}{*}{Persons} & Up-to-1hop & 0.64 (C) & 0.32 (P) & -- \\
 &  & Up-to-2hop & 0.00 (C) & 0.56 (C) & -- \\
\cmidrule(lr){2-6}
 & \multirow{2}{*}{Organizations} & Up-to-1hop & 0.53 (C) & 0.63 (P) & -- \\
 &  & Up-to-2hop & 0.00 (C) & 0.43 (C) & -- \\
\cmidrule(lr){2-6}
 & \multirow{2}{*}{Organisms} & Up-to-1hop & 0.85 (C) & 0.43 (P) & -- \\
 &  & Up-to-2hop & 0.11 (C) & 0.74 (C) & -- \\
\midrule
\multirow{12}{*}{Llama 3.1 8B} & \multirow{3}{*}{Locations} & Up-to-1hop & 0.41 (C) & 0.81 (P) & 0.77 (P) \\
 &  & Up-to-2hop & 0.20 (C) & 0.40 (C) & 0.80 (P) \\
 &  & Up-to-3hop & 0.14 (C) & 0.33 (C) & 0.37 (C) \\
\cmidrule(lr){2-6}
 & \multirow{3}{*}{Research Topic} & Up-to-1hop & 0.04 (C) & 0.79 (P) & 0.72 (P) \\
 &  & Up-to-2hop & 0.00 (C) & 0.10 (C) & 0.72 (P) \\
 &  & Up-to-3hop & 0.00 (C) & 0.05 (C) & 0.07 (C) \\
\cmidrule(lr){2-6}
 & \multirow{2}{*}{Persons} & Up-to-1hop & 0.68 (C) & 0.93 (P) & -- \\
 &  & Up-to-2hop & 0.00 (C) & 0.54 (C) & -- \\
\cmidrule(lr){2-6}
 & \multirow{2}{*}{Organizations} & Up-to-1hop & 0.62 (C) & 1.00 (P) & -- \\
 &  & Up-to-2hop & 0.00 (C) & 0.17 (C) & -- \\
\cmidrule(lr){2-6}
 & \multirow{2}{*}{Organisms} & Up-to-1hop & 0.82 (C) & 0.93 (P) & -- \\
 &  & Up-to-2hop & 0.27 (C) & 0.65 (C) & -- \\
\midrule
\multirow{12}{*}{Qwen3 8B} & \multirow{3}{*}{Locations} & Up-to-1hop & 0.34 (C) & 0.71 (P) & 0.73 (P) \\
 &  & Up-to-2hop & 0.01 (C) & 0.28 (C) & 0.73 (P) \\
 &  & Up-to-3hop & 0.01 (C) & 0.24 (C) & 0.32 (C) \\
\cmidrule(lr){2-6}
 & \multirow{3}{*}{Research Topic} & Up-to-1hop & 0.05 (C) & 0.82 (P) & 0.90 (P) \\
 &  & Up-to-2hop & 0.00 (C) & 0.06 (C) & 0.89 (P) \\
 &  & Up-to-3hop & 0.00 (C) & 0.01 (C) & 0.03 (C) \\
\cmidrule(lr){2-6}
 & \multirow{2}{*}{Persons} & Up-to-1hop & 0.53 (C) & 0.83 (P) & -- \\
 &  & Up-to-2hop & 0.00 (C) & 0.50 (C) & -- \\
\cmidrule(lr){2-6}
 & \multirow{2}{*}{Organizations} & Up-to-1hop & 0.40 (C) & 1.00 (P) & -- \\
 &  & Up-to-2hop & 0.03 (C) & 0.26 (C) & -- \\
\cmidrule(lr){2-6}
 & \multirow{2}{*}{Organisms} & Up-to-1hop & 0.67 (C) & 0.78 (P) & -- \\
 &  & Up-to-2hop & 0.00 (C) & 0.71 (C) & -- \\
\midrule
\multirow{12}{*}{Qwen3 14B} & \multirow{3}{*}{Locations} & Up-to-1hop & 0.27 (C) & 0.66 (P) & 0.62 (P) \\
 &  & Up-to-2hop & 0.00 (C) & 0.24 (C) & 0.62 (P) \\
 &  & Up-to-3hop & 0.00 (C) & 0.22 (C) & 0.31 (C) \\
\cmidrule(lr){2-6}
 & \multirow{3}{*}{Research Topic} & Up-to-1hop & 0.02 (C) & 0.84 (P) & 0.85 (P) \\
 &  & Up-to-2hop & 0.00 (C) & 0.04 (C) & 0.85 (P) \\
 &  & Up-to-3hop & 0.00 (C) & 0.02 (C) & 0.07 (C) \\
\cmidrule(lr){2-6}
 & \multirow{2}{*}{Persons} & Up-to-1hop & 0.37 (C) & 0.73 (P) & -- \\
 &  & Up-to-2hop & 0.00 (C) & 0.30 (C) & -- \\
\cmidrule(lr){2-6}
 & \multirow{2}{*}{Organizations} & Up-to-1hop & 0.37 (C) & 1.00 (P) & -- \\
 &  & Up-to-2hop & 0.00 (C) & 0.09 (C) & -- \\
\cmidrule(lr){2-6}
 & \multirow{2}{*}{Organisms} & Up-to-1hop & 0.52 (C) & 0.68 (P) & -- \\
 &  & Up-to-2hop & 0.02 (C) & 0.58 (C) & -- \\
}

\let\cfBaselineTable\relax

\end{document}